# Deep Learning Based Virtual Point Tracking for Real-Time Target-less Dynamic Displacement Measurement in Railway Applications

Dachuan Shi [*a], Eldar Šabanovič[b], Luca Rizzetto[c], Viktor Skrickij[b], Roberto Oliverio[c], Nadia Kaviani[c], Yunguang Ye [a], Gintautas Bureika[b], Stefano Ricci[c], Markus Hecht[a]

[a] *Institute of Land and Sea Transport Systems, Technical University of Berlin, Berlin 10587, Germany*
[b] *Faculty of Transport Engineering, Vilnius Gediminas Technical University, LT-10223 Vilnius*
[c] *Department of Buildings and Environmental Engineering, Sapienza University of Rome, 00185 Roma*

* Corresponding E-mail: dachuan.shi@tu-berlin.de, Tel.: +49 030 314 79806 and Fax: +49 030 314 22529

## Abstract

In the application of computer-vision-based displacement measurement, an optical target is usually required to prove the reference. If the optical target cannot be attached to the measuring objective, edge detection, feature matching, and template matching are the most common approaches in target-less photogrammetry. However, their performance significantly relies on parameter settings. This becomes problematic in dynamic scenes where complicated background texture exists and varies over time. We propose virtual point tracking for real-time target-less dynamic displacement measurement, incorporating deep learning techniques and domain knowledge to tackle this issue. Our approach consists of three steps: 1) automatic calibration for detection of region of interest; 2) virtual point detection for each video frame using deep convolutional neural network; 3) domain-knowledge based rule engine for point tracking in adjacent frames. The proposed approach can be executed on an edge computer in a real-time manner (i.e. over 30 frames per second). We demonstrate our approach for a railway application, where the lateral displacement of the wheel on the rail is measured during operation. We also implemented an algorithm using template matching and line detection as the baseline for comparison. The numerical experiments have been performed to evaluate our approach's performance and latency in a harsh railway environment with dynamic complex backgrounds. We make our code and data available at https://github.com/quickhdsdc/Point-Tracking-for-Displacement-Measurement-in-Railway-Applications.

*Keywords*: Point tracking; Computer vision; Displacement measurement; Photogrammetry; Deep learning; Railway

## 1. Introduction

### 1.1. Background and motivation

Thanks to the rapid advance in computer vision (CV) in the last decade, there is a noticeable increase in many sectors applying photogrammetry to inspect structures. A typical photogrammetry application is the deformation measurement of large structures such as bridges in civil engineering [1]. In the railway sector, Zhan et al. [2] proposed to use high-speed line scan cameras to measure catenary geometry parameters, calibrated by a 1-D optical target. Li et al. [3] used CV to monitor track slab deformation. Two optical targets were attached to the track slab to extract region of interest (RoI). In the aforementioned applications, optical targets are required to provide measurement references. When optical targets cannot be attached to the structure, edge detection, digital image correlation, template matching and template matching are the most common solutions [4][5]. However, they may suffer from robustness problems due to complex backgrounds. Wang et al. [6] combined a deep learning model and a template-matching driven tracking algorithm for recognition and tracking of rail profile from laser fringe images. Jiang et al. [7] proposed a robust line detection workflow for the uplift measurement of railway catenary, addressing the problem caused by noisy background. The measurement was done in a static condition by fixing the camera system next to the railway. The challenge we are facing is more complex. We are addressing the issue of real-time target-less dynamic displacement measurement in front of noisy and varying backgrounds. In the context of the railway, we aim to monitor the wheels' lateral motion of a railway vehicle relative to the rail in regular railway operation. It tackles an unsolved railway issue related to track geometry (TG) monitoring .

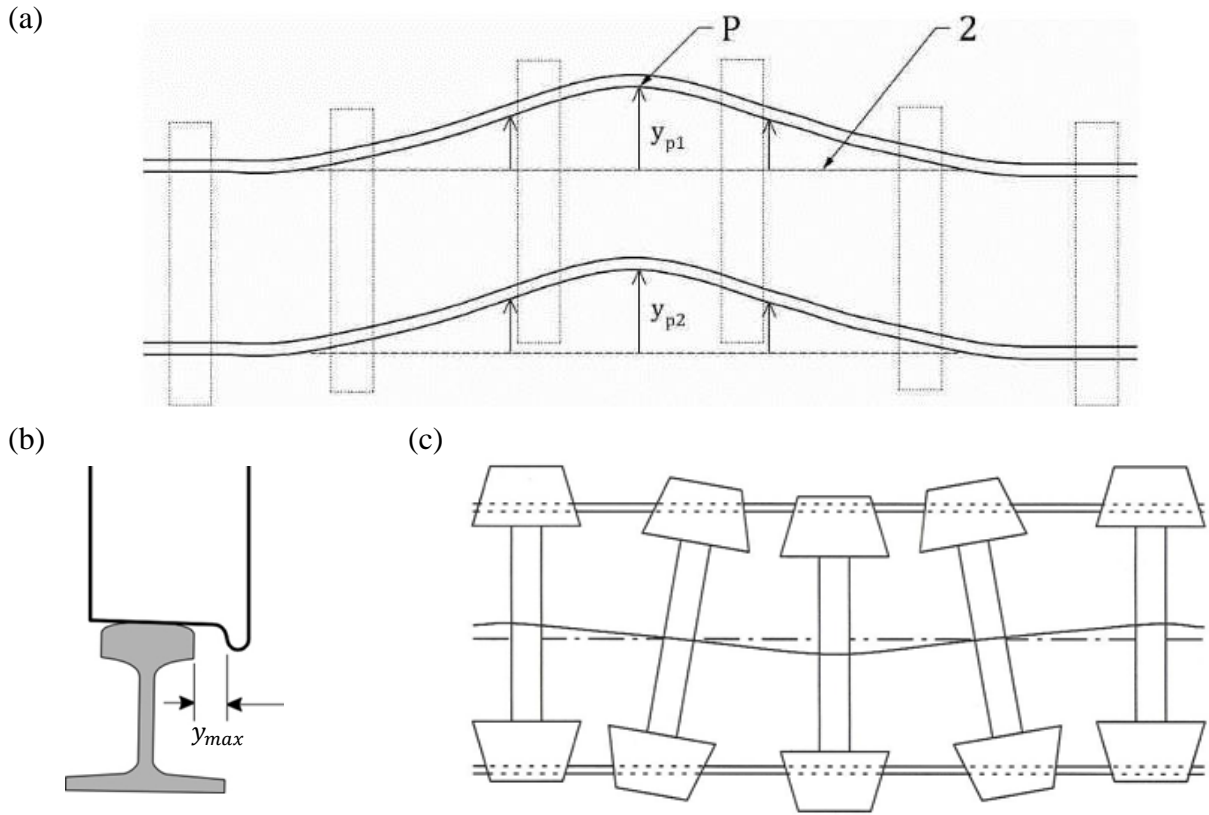

**Fig. 1.** (a) lateral alignment of the left rail $y_{p1}$ and right rail $y_{p2}$, where P denotes the rail reference point and "2" denotes the reference rail line [8]; (b) illustration of wheel/rail gauge clearance [9]; (c) schematic view of hunting motion [9]

TG parameters are defined as the indicators for track maintenance in the European railway standard EN 13848-1 [8], namely track gauge, cross-level, longitudinal level, lateral alignment, twist. Based on the statistical study in a European project [10], longitudinal level and lateral alignment are the most critical parameters for maintenance decisions. Other parameters are either highly linearly correlated to longitudinal level or degrading slower than longitudinal level. Traditionally, TG parameters are measured by the dedicated TG inspection systems in regular inspections, which are typically based on high-value laser triangulation sensors or/and inertial measurement units (IMU). The inspection interval is usually defined as several months, which results in the lack of up-to-date information on track conditions [11]. In order to improve the information availability and enable efficient maintenance decisions, TG monitoring on in-service vehicles was proposed. TG monitoring has been studied extensively in the last two decades [12]. Accelerometers have been commonly accepted as the most promising sensor for TG monitoring due to their low cost and robustness. It has been validated in previous studies that longitudinal level can be accurately reconstructed from vertical accelerations [12]. However, lateral alignment cannot be accurately derived from lateral accelerations due to railway vehicle dynamics. As shown in Fig. 1 (a), lateral alignment is defined as the lateral deviation $y_p$ between the actual and reference rail line in the horizontal plane at the point $P$ on the each rail, being at the position 14 mm below the top of the railhead for the standard rail profile UIC 60E1 [8]. It is expected that the vehicle wheels follow the excitation of lateral alignment in the lateral direction so that lateral alignment can be estimated by accelerations. However, the wheels do not follow lateral alignment exactly as the vertical one. One reason is that the wheel has a freedom of movement in the lateral direction in a clearance $y_{max}$, which refers to the clearance between the wheel flange and the rail head edge, as shown in Fig. 1 (b). Another reason given by True et al. [13] is that the lateral irregularities simultaneously act on the lateral force and the spin torque of the wheel-rail contact force, which are nonlinearly coupled. This indicates that lateral alignment cannot be accurately derived from the perspective of vehicle dynamics.

To tackle this issue, Ripke et al. [10] combined acceleration measurements with a multi-body dynamic simulation (MBS) model of the vehicle, on which the accelerometers were installed. The alignment was estimated by accelerations and then corrected by the MBS model using a dedicated correction mechanism. The estimated alignment was compared with the one measured by a commercial TG inspection system. However, this approach was vitiated by the comparison results. Rosa et al. [14] proposed a model-based method, combing MBS and Kalman filter, to estimate lateral alignment. However, a critical issue of a model-based method is that the model cannot take into account the wear process of wheel profile, which has significant effects on vehicle dynamics. Rosa et al. [15] proposed to train a machine learning (ML) based classifier to detect large track lateral irregularities. From a maintenance perspective, two classes of alignment values have been defined by thresholding. The measured alignment values in class 1 indicate the normal track condition, and no specific maintenance measure has to be taken. Class 2 indicates severe track degradation, requiring short-term maintenance measures. As well known, the features as the input for the classifier are essential for classification performance. In [15], only standard deviations of accelerations were defined as the features, which may not contain abundant classification information. The test accuracy was under 90%. This approach also evaded the reconstruction of alignment values.

Based on the previous studies, we conclude that the wheels' lateral displacement on the rail (LDWR) is indispensable to estimate the accurate lateral alignment. Therefore, we propose deep-learning (DL) based virtual point tracking to measure LDWR in a real-time manner. Combined with an accelerometer, the proposed system can be used to reconstruct the alignment for a massive deployment on in-service trains.

Our approach can also be used for hunting detection, as shown in Fig. 1 (c), which indicates the dynamic instability of railway vehicles and is thus safety-relevant. The current detection methods are based on acceleration measurements. The detection performance could be interfered by alignment, particularly when detecting small-amplitude hunting instability [16]. Monitoring LDWR can fundamentally solve this problem. Furthermore, monitoring LDWR is a central part of the active wheelset steering systems using lateral displacement control strategy [17]. LDWR can express the rolling radius of the wheels. If the lateral displacement can satisfy a specific condition, the wheelset will be in pure rolling condition, resulting in minimal wear in a curve. Within the control chain, the measured LDWR provides feedback to the control system [16].

### 1.2. Related work

Our task is to detect and track the virtual points for target-less dynamic displacement measurement in front of noisy and varying backgrounds. We introduce DL approaches for human pose estimation (HPE) for point detection. In the following, we review the related work for the measurement of LDWR, photogrammetry for displacement measurement and DL-based HPE, respectively.

A commercial system based on laser triangulation sensors was used to measure LDWR for active wheel control [18]. The laser sensors were mounted on the wheelset axle, closely pointing at the railhead. The accuracy of the laser sensors is of the order of 0.1 mm. However, the sensors are subject to high vibrations at the wheelset level, which could degrade their lifetime and performance. Kim [17] used a charge-coupled device (CCD) camera to measure LDWR for active wheel control. LDWR was measured by tracking the rail line and the wheel template. The proposed algorithm was mainly based on conventional image processing techniques of filtering, edge detection, template matching and line detection. However, it requires parameter tuning as a part of calibration for different environmental conditions, which is a laborious and time-consuming process. Yamamoto [19] used a thermographic camera installed on the bogie frame to view the wheel-rail contact area. Despite successful localization, the thermographic camera has a low resolution of 320×240 pixels and thus a low measurement resolution in millimeters. It cannot fulfill the requirements of TG monitoring.

Photogrammetry for displacement measurement is typically divided into five steps: camera calibration, RoI selection, feature extraction, visual tracking and displacement calculation. The applicable methods for each step have been reviewed by Baqersad et al. [4] and Dong et al. [5]. Edge detection and template matching algorithms are frequently applied in target-less photogrammetry, where structures' inner edges or features are extracted for object detection and tracking. Guo et al. [20] introduced Lucas-Kanade template tracking algorithm for dynamic displacement measurement. This algorithm was able to process images from high-speed cameras. However, it requires a pre-defined template that remains visually stable within the measurement. This prerequisite may not be fulfilled in the case of noisy and dynamic backgrounds. Cha et al. [21] applied a phased-based optical flow algorithm for motion tracking of structures. However, optical flow approaches are sensitive to the variation of illumination and backgrounds. Dong et al. [22] applied spatio-temporal context learning for RoI tracking and Taylor approximation for subpixel motion estimation. The robustness of this approach has been validated for small motion tracking in laboratory experiments by varying the illumination and humidity. Apart from the conventional image processing techniques, DL has been introduced in photogrammetry. Yang et al. [23] combined convolutional neural network (CNN) and recurrent neural network (RNN) for modal analysis of the structures. A vanilla CNN model was used for spatial feature extraction, while a long short-term memory (LSTM) network was used to model the temporal dependency over the measurement period. The outputs were the natural frequencies. In the images, the specimens were highlighted through the laser point of a laser vibrometer, which was intended to provide the ground-truth natural frequencies. This laser point may unexpectedly become the optical target and lead to success. However, this was not analyzed in the paper. Liu et al. [24] used CNN for vibration frequency measurement of bridges. The 9×9 RoI in the frames was manually selected and flattened as 1-D sequences fed into CNN as the inputs. The CNN outputted vibration frequencies. The manual selection of RoI played an essential role. RoI must contain an objective with clear edges and a clear background.

Displacement could be measured by tracking reference points, which conventionally refer to optical targets. Alternatively, virtual points can be defined in measuring objectives and automatically detected by employing advanced CV techniques. A successful application of virtual point detection/tracking is HPE. HPE is a fundamental CV task, aiming to estimate the posture of human bodies. In the last decade, CV-based HPE has been under rapid development thanks to DL techniques [25]. For HPE, the virtual points are defined as a series of points at a human body's kinematic joints [26], such as eyes, neck, elbows and ankles. In terms of problem formulation, the methods for 2D HPE fall into two categories, namely regression-based and detection-based methods [25]. Detection-based methods transfer the virtual points into 2D representations (e.g. heatmaps) and then map the input image to these heatmaps. This method is commonly used in the modern CNN architectures for HPE, such as the stacked hourglass network [27], the encoder-decoder network [28] and the high-resolution network [29]. In contrast, regression-based methods directly output the coordinates of the virtual points from a given image. It is much harder to map the input 2D image directly to the point coordinates than to the 2D heatmaps. Therefore, a more powerful backbone architecture is required. The CNN network architecture proposed by Luvizon et al. [30] consisted of Inception-V4 for feature extraction and multiple prediction blocks to predict the heatmap of each point. Finally, the Soft-argmax layer was added to regress the coordinates of a keypoint from the heatmap. In recent work, Bazarevsky et al. [31] combined both methods in one network. The network has two heads in the training process, one for prediction of the heatmap and the other for regression of the coordinates. Only the regression head is kept for online inference, while the heatmap prediction head is removed.

*1.3. Challenges and contributions*

In our railway application for dynamic displacement measurement, we are facing the following challenges. Firstly, it is a monitoring task, rather than an inspection. Monitoring devices are typically developed for the massive deployment and full automation during operation. Therefore, Monitoring devices are expected to have high automation and low investment costs. Secondly, the CV system is installed on the railway vehicle facing a wheel, moving along the railway track. An optical target cannot be attached to the rotating wheel. The common target-less approaches, such as edge detection, template matching and line detection, are prone to performance losses in front of dynamic complex backgrounds, where complex textures such as ballast, sleepers and plants exist and vary over time. Thirdly, the images should be processed in a real-time manner, as the calculated LDWR has to be fused with the acceleration measurements to reconstruct track lateral alignment. To address these challenges, we propose a novel approach to virtual point tracking. To our best knowledge, our work is the first attempt to combine HPE and domain knowledge for displacement measurement.

In this paper, we mainly focus on the proposed algorithm for virtual point tracking. The calculation of displacement between the virtual points has been introduced and validated in [32]. The fusion of CV and accelerometers will be addressed in future work. Our main contributions are summarized as follow:

1) A novel approach of virtual point tracking for target-less displacement measurement is proposed, consisting of RoI detection, point detection and point tracking.
2) A lightweight CNN architecture is proposed for real-time point detection in each video frame.
3) A rule engine based on railway domain knowledge is defined for point tracking.
4) Implementation of the proposed approach for real-time edge computing.
5) We make our code and data available at https://github.com/quickhdsdc/Point-Tracking-for-Displacement-Measurement-in-Railway-Applications

The structure of the paper is as follows. Section 2 briefly introduces the hardware of the designed monitoring system. Section 3 describes the proposed approach for virtual point tracking in detail, the implemented baseline method, and the image corruption methods for data augmentation. In Section 4, extensive experiments are conducted to evaluate as well as validate each step in our approach and demonstrate the entire approach. In addition, computational complexity and robustness are discussed. Section 5 draws the conclusions.

## 2. Hardware components of the monitoring system

The proposed monitoring system consists of an off-the-shelf stereo camera, an air cleaning system, a processing unit, a lighting system, and a mounting system with dampers. The air cleaning system aims to clean the camera lens by blowing compressed air regularly. This is a standard solution to avoid dirt in the optical systems [12]. From the software perspective, we enhance the robustness of the algorithm against the image's visual corruptions. This will be introduced in Section 4. For optical sensing, ZED2 stereo camera is used in our system [33], which is configured to output videos with the resolution of 1920×1080 pixels at the sample rate of 30 frames per second (fps). Any comparable cameras can be used as well. The depth information is merely used for displacement measurement. The algorithms described in this are directly applicable for 2D images obtained by regular CCD cameras. As the processing unit, Nvidia Jetson Tx2 has 256 core NVIDIA Pascal architecture and ARMv8 6 core multi-processor CPU complex, enabling real-time execution of DL models [34]. The mounting system consisting of the vibration dampers, a crossbar and a clamp can be easily installed on different bogies types. The camera housing is equipped with an external lighting system, which consists of a series of LEDs. The entire system is installed on the bottom of the bogie frame, facing the wheel, as shown in Fig. 2. Two systems are required to monitor the wheel-rail pair on the left and right sides simultaneously. In the current hardware implementation, the cleaning system is not included. The processing unit is inside the vehicle cabin, connecting to and powering the camera. The hardware of the monitoring system will be further improved for long-term monitoring.

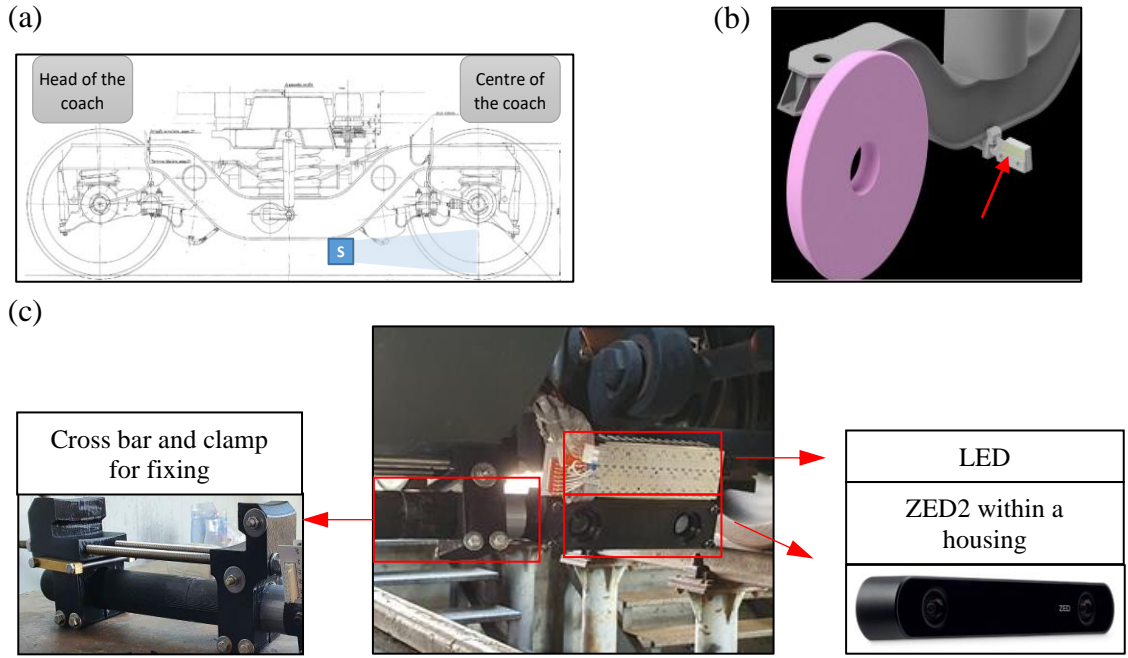

**Fig. 2.** (a) drawing of the camera position on the bogie frame; (b) CAD model of the camera installation; (c) monitoring system installed on the bogie frame

## 3. Approach for virtual point tracking

We formulate the task of dynamic displacement measurement to track virtual reference points and calculate the distance between two virtual reference points. This paper focuses on virtual point tracking. The displacement calculation method has been introduced and validated in our previous work [32]. We define three reference points on the wheel ($P_w$) and rail ($P_{r1}$ and $P_{r2}$) respectively. $P_{r1}$ refers to the reference point $P$ for lateral alignment [8], as introduced in Section 1.1. $P_{r2}$ is the symmetry point of $P_{r1}$ on the other side of the railhead edge. The distance between $P_{r1}$ and $P_{r2}$ is the width of the railhead. The lateral displacement $D$ of the wheel on the rail (LDWR) is represented by the lateral distance between $P_w$ and $P_{r1}$, see Fig. 3. The relative lateral motion of the wheel is represented by the changes of $D$ (i.e. $\Delta D$) over time, which is the output of the monitoring system. The point $P_{r2}$ is defined for tracking mechanism, which is explained in Section 3.3.

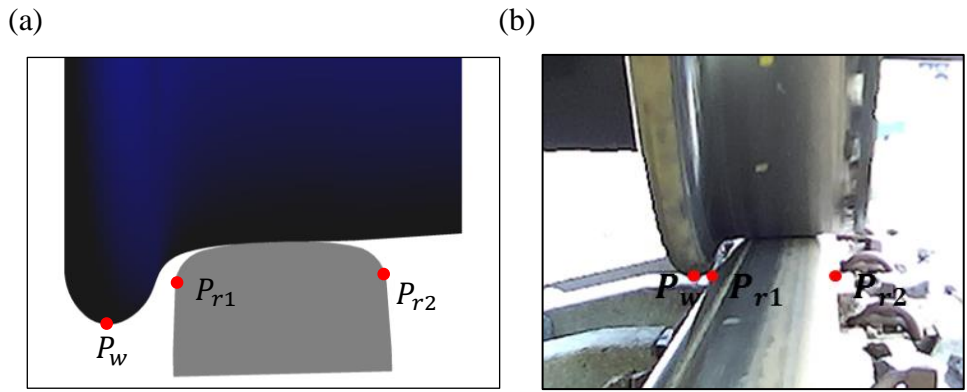

**Fig. 3.** (a) defined keypoints $P_w$, $P_{r1}$ and $P_{r2}$ illustrated in the animation; (b) marked in the real photo (right)

Virtual point tracking consists of three steps, as shown in Fig. 4. The first step is the calibration, executed for the first-time installation. This calibration process detects RoI, which refers to the wheel-rail contact area.

The outputs are the coordinates of the centre point of RoI. Moreover, the distance between the camera and the wheel is obtained as the stereo camera's depth. The distance is an input parameter for displacement calculation. In the case of using a CCD camera, the distance has to be manually measured. The next steps are executed to detect and track virtual points in real-time. Next, we will introduce each step in detail.

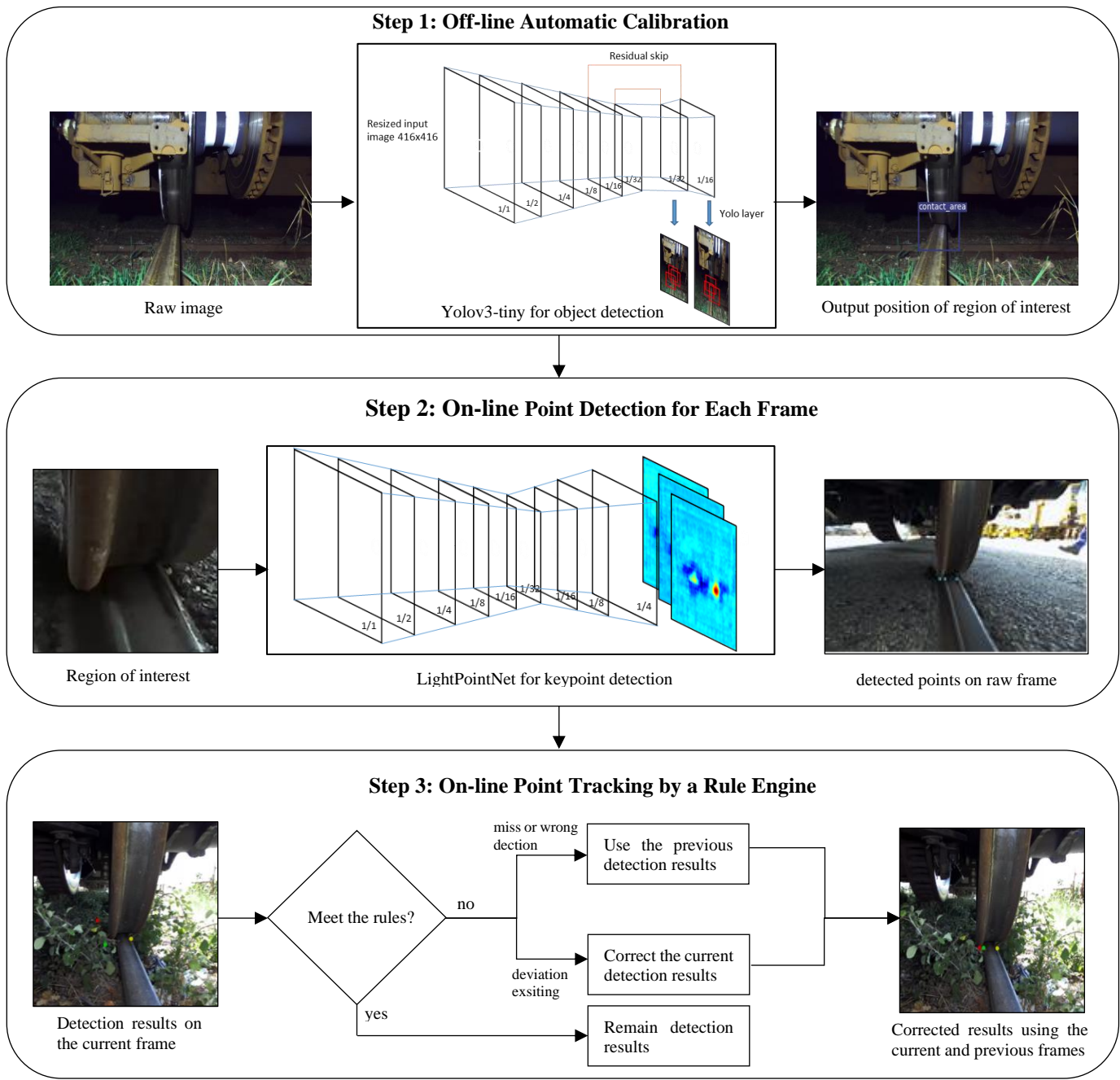

**Fig. 4.** Approach to track virtual points on the wheel and rail

### 3.1. Step 1: Off-line automatic calibration

As the resolution of each frame is 1920×1080 pixels, it is necessary to resize the image prior to feeding it to CNN. However, resizing and restoring the image cause the additional measurement error for point detection. To avoid the step of image resizing, we propose cropping the RoI from the raw image. We choose a mature object detection technique based YOLOv3 [35], which is a CNN architecture and has been widely deployed for diverse applications. We adopt a modified version of YOLOv3 for RoI detection, called YOLOv3-tiny. The architecture of YOLOv3-tiny is shown in Fig. 4 and Tab. 1. The first 13 layers are used for feature extraction, known as Darknet. The input for Darknet is the images with 416×416 pixels downsized from the original 1920×1080 ones. The output 1024 feature maps of Darknet have the dimension of $13 \times 13$ pixels. The layers 13-16 are to make predictions. YOLOv3-tiny pre-defines 3 anchor boxes to predict the objects' bounding boxes and generates 6 parameters for each bounding box, i.e. x and y coordinate of its center point, its width and height, a prediction score, and the object class. In our case, only one class is defined for the wheel-rail contact area, while others are backgrounds. The 14 and 15 layers downsized the number of the feature maps to 18. Each $13 \times 13$ feature map regresses one parameter. The 16th layer compares the prediction with the ground truth to calculate the loss. The loss consists of classification loss, localization loss, and confidence loss. The detailed loss functions can be found in the original paper [35]. YOLOv3-tiny predicts at two different scales. The first scale uses the aforementioned $13 \times 13$ feature maps and passes the feature maps to the second scale, which is implemented in the layers 17-23. The outputs of YOLOv3-tiny are the candidates of RoI with the dimension of $N \times 18$, where $N$ denotes the number of the candidates. The final prediction is selected by objectness score thresholding and non-maximal suppression [35].

**Tab. 1.** Adapted YOLOv3-tiny architecture

| Layer Index | Type | Filters | Size/Stride | Output |
|---|---|---|---|---|
| 0 | Convolutional | 16 | $3 \times 3/1$ | $416 \times 416$ |
| 1 | Maxpool | | $2 \times 2/2$ | $208 \times 208$ |
| 2 | Convolutional | 32 | $3 \times 3/1$ | $208 \times 208$ |
| 3 | Maxpool | | $2 \times 2/2$ | $104 \times 104$ |
| 4 | Convolutional | 64 | $3 \times 3/1$ | $104 \times 104$ |
| 5 | Maxpool | | $2 \times 2/2$ | $52 \times 52$ |
| 6 | Convolutional | 128 | $3 \times 3/1$ | $52 \times 52$ |
| 7 | Maxpool | | $2 \times 2/2$ | $26 \times 26$ |
| 8 | Convolutional | 256 | $3 \times 3/1$ | $26 \times 26$ |
| 9 | Maxpool | | $2 \times 2/2$ | $13 \times 13$ |
| 10 | Convolutional | 512 | $3 \times 3/1$ | $13 \times 13$ |
| 11 | Maxpool | | $2 \times 2/1$ | $13 \times 13$ |
| 12 | Convolutional | 1024 | $3 \times 3/1$ | $13 \times 13$ |
| 13 | Convolutional | 256 | $1 \times 1/1$ | $13 \times 13$ |
| 14 | Convolutional | 512 | $3 \times 3/1$ | $13 \times 13$ |
| 15 | Convolutional | 18 | $1\times 1/1$ | $13 \times 13$ |
| 16 | YOLO | | | |
| 17 | Route 13 | | | |
| 18 | Convolutional | 128 | $1\times 1/1$ | $13 \times 13$ |
| 19 | Upsampling | | $2 \times 2/1$ | $26 \times 26$ |
| 20 | Route 19, 8 | | | |
| 21 | Convolutional | 256 | $3 \times 3/1$ | $26 \times 26$ |
| 22 | Convolutional | 18 | $1\times 1/1$ | $26 \times 26$ |
| 23 | YOLO | | | |

### 3.2. Step 2: On-line point detection for each frame

Point detection is an essential step in our approach. We propose LightPointNet, a lightweight CNN architecture using integral regression for real-time point detection on each video frame. LightPointNet consists of an encoder for hierarchical feature extraction and a decoder for heatmap prediction. Inspired by [28,36,37], the key insights behind LightPointNet are the lightweight backbone, the straightforward encoder-decoder structure and integral loss.

The architecture of LightPointNet is shown in Fig. 4 and Tab. 2. The first 12 blocks build the encoder, while the last four blocks build the decoder. The whole network is built by stacking three building blocks. The first block "Conv" refers to a convolutional block, consisting of a convolutional layer, a batch normalization layer and the hard-swish function (HS) as the activation function for nonlinearity (NL). In this block, 16 convolutional filters parameterized by the weights $W \in \mathcal{R}^{3\times3}$ are performed on the input image $I \in \mathcal{R}^{256 \times 256 \times 3}$ to generate the feature map $F \in \mathcal{R}^{128 \times 128 \times 16}$. Then, mini-batch normalization [38] and hard-swish [36] are performed on the feature map $F$ to reduce internal corvariate shift and add nonlinearity. The swish function aims to solve the dead neuron problem of ReLu, which is the most common activation function for CNN. The hard version of the swish function reduces the computational complexity of the original one, defined as:

$$Hswish = x \cdot (ReLu6(x + 3)) \,/\, 6 \tag{1}$$

$$ReLu6 = min \,(\max\,(0, x), 6) \tag{2}$$

**Tab. 2.** LightPointNet architecture

| Block | Type | NL | SE | Exp size | Filters | Size/Stride | Output |
|-------|------|-----|-------|----------|---------|-------------|--------|
| 0 | Conv | HS | false | - | 16 | $3 \times 3/2$ | $128 \times 128 \times 16$ |
| 1 | Bneck | RE | false | 16 | 16 | $3 \times 3/2$ | $64 \times 64 \times 16$ |
| 2 | Bneck | RE | true | 72 | 24 | $3 \times 3/2$ | $32 \times 32 \times 24$ |
| 3 | Bneck | RE | false | 88 | 24 | $3 \times 3/1$ | $32 \times 32 \times 24$ |
| 4 | Bneck | HS | false | 96 | 40 | $5 \times 5/2$ | $16 \times 16 \times 40$ |
| 5 | Bneck | HS | true | 240 | 40 | $5 \times 5/1$ | $16 \times 16 \times 40$ |
| 6 | Bneck | HS | true | 240 | 40 | $5 \times 5/1$ | $16 \times 16 \times 40$ |
| 7 | Bneck | HS | true | 120 | 48 | $5 \times 5/1$ | $16 \times 16 \times 48$ |
| 8 | Bneck | HS | true | 144 | 48 | $5 \times 5/1$ | $16 \times 16 \times 48$ |
| 9 | Bneck | HS | true | 192 | 64 | $5 \times 5/2$ | $8 \times 8 \times 64$ |
| 10 | Bneck | HS | true | 384 | 64 | $5 \times 5/1$ | $8 \times 8 \times 64$ |
| 11 | Bneck | HS | true | 384 | 64 | $5 \times 5/1$ | $8 \times 8 \times 64$ |
| 12 | ConvTranspose | RE | false | - | 256 | $4 \times 4/2$ | $16 \times 16 \times 256$ |
| 13 | ConvTranspose | RE | false | - | 256 | $4 \times 4/2$ | $32 \times 32 \times 256$ |
| 14 | ConvTranspose | RE | false | - | 256 | $4 \times 4/2$ | $64 \times 64 \times 256$ |
| 15 | Conv | - | false | - | 3 | $1 \times 1/1$ | $64 \times 64 \times 3$ |

The convolutional block is followed by 11 blocks of inverted residual and linear bottleneck (Bneck) [36]. Bneck is a modified version of the original residual operation [39], which enables the skip connection between the input and output feature maps by following a wide-narrow-wide bottleneck structure in terms of the channel number. Bneck uses an inverted bottleneck with a narrow-wide-narrow structure. It is implemented by stacking three convolutional layers. The first one is $1 \times 1$ pointwise convolution to expand the input channel dimension $c$ by a factor $e$, followed by an activation function. The expanded size for each Bneck block is given in the column "Exp size" of Tab. 2. The second one is $3 \times 3$ depthwise convolution with an activation function, keeping the channel dimension unchanged. Replacing regular convolution with depthwise convolution is an effective lightweight measure. This will be further described in Section 4.6. The third one is $1 \times 1$ pointwise convolution to recover the output channel dimension to $c$, allowing the identity skip connection between the block inputs and outputs. The third convolution layer does not involve an activation function and thus remains linear. Bneck can be combined with Squeeze-and-Excite (SE) [40], which improves channel interdependencies of feature maps. The column "SE" indicates the presence of the SE module. The structure of SE can be found in [40]. We merely replace the original sigmoid activation with the hard sigmoid function, which is defined as:

$$HSig = ReLu6(x + 3) \tag{3}$$

The Bneck blocks extract hierarchical features and downsize the feature maps to $8 \times 8$. Afterward, 3 blocks of "ConvTranspose" are stacked to upsample the feature maps to $64 \times 64$. ConvTranspose consists of a transposed convolutional layer, a batch normalization layer, and an activation function. The final Conv block aims to output the final heatmaps $H \in \mathcal{R}^{64 \times 64 \times 3}$ for the defined three virtual points $P_w$, $P_{r1}$ and $P_{r2}$, respectively.

The common regression-based HPE methods use mean square errors (MSE) between the predicted heatmaps and the ground-truth ones as the optimization objective for CNN training. The point coordinates are obtained using arguments of the maxima (argmax). This induces inevitable quantization errors. In our case, the resolution of the heatmaps is one-fourth of the input image. That means the quantization errors and the prediction errors at the heatmap level are enlarged four times. To avoid this, we use the discrete softmax function instead of argmax, which is differentiable and thus can be included in the training process [37]. The predicted point coordinates are the integration of the heatmap weights in all locations along the x and y-axis, respectively. In this way, the predicted point coordinates are continuous and thus on a subpixel level. Fig. 5 exemplarily shows the prediction process of a wheel reference point. Finally, the loss function as the training objective, termed integral loss, is defined as the absolute differences between the ground-truth coordinates and the predicted ones using integral regression.

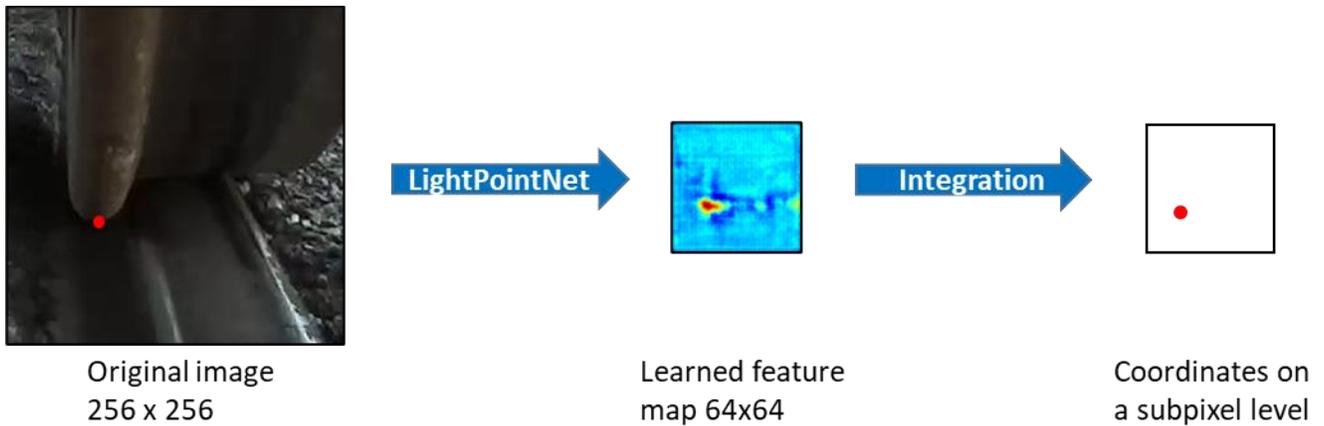

**Fig. 5.** Prediction process of a wheel reference point

### 3.3. Step 3: On-line point tracking by a rule engine

LightPointNet may output false detections during regular railway operation, especially in corner cases. For instance, as shown in the third block in Fig. 4, the grass occludes the points $P_w$ and $P_{r1}$. A correct point detection is impossible on this single frame. To correct false detections, we propose a rule engine as the point tracker. The rule machine independent from CNN has two advantages in our application. Unlike the problems of object tracking and human pose tracking, our railway application has similar scenes, i.e. the wheels running on the rails. The virtual points have spatial correlations with each other under specific geometric constraints. This allows defining the rules based on railway domain knowledge. On the other hand, we have specific challenges in terms of data availability, which is a common issue for any domain-specific application. As in a recent work of real-time human pose tracking indicated [31], 85k annotated images were used to train a pose tracking network. In industrial practice, data collection and annotation are laborious and costly. Much fewer data obtained in field tests are available to train CNN. Therefore, we combine a DL-based point detector with a domain-knowledge-based tracker to achieve real-time point tracking, which requires much less training data. Furthermore, the rule machine can automatically identify the corner cases, once the CV system is deployed for a long-term trial. The corresponding video frames can be collected to update the model for performance improvement.

The flow chart of the rule engine is shown in Fig. 4 Step 3. We define the following rules as well as the corresponding indicators and thresholds in Tab. 3. Each rule is independently examined. Rule 1 and 2 constrain the y-coordinates of the virtual points, which represent the projection of the relative vertical and longitudinal motion between the camera and the wheel in the horizontal plane. Three virtual points are defined at the same horizontal level, i.e. $y_{ref}$. The relative movement of rail reference points $P_{r1}$ and $P_{r2}$ does not exist. The only reasonable disparity of y-coordinates between the wheel and rail reference point is linked with wheel bounce due to a high excitation of rail irregularities. However, this is a rare event and can be compensated by wheel acceleration measurement. Therefore, we consider that the y-coordinates of three points should vary by a small margin. Root mean squared error (RMSE) is used as the indicator for Rule 1 and 2. When $\sigma_y$ in Rule 1 exceeds the threshold $\sigma_{yTH_0}$, the detection results are regarded as unreliable. The detection results for the current frame are thus inherited from the previous frame. When $\sigma_y$ lies between $\sigma_{yTH_0}$ and $\sigma_{yTH_1}$, a correction mechanism is applied to the detection results. We take the averaged coordinates of the previous and current frames as the corrected values. The values of the threshold $\sigma_{yTH_0}$ and $\sigma_{yTH_1}$ indicate the error tolerance of the virtual point detection results. As the detection errors are expected to be at the level of 1 mm, we empirically define $\sigma_{yTH_0} = 5$ and $\sigma_{yTH_1} = 10$. The thresholds do not have to be changed when the monitoring system is installed on a different vehicle. Similarly, Rule 3 and 4 constrain the difference of x-coordinates between $P_{r1}$ and $P_{r2}$, as it represents the railhead's width. In practice, the rail head width may vary at a small margin due to wear. Rule 5 constrains the difference of x-coordinates between $P_w$ and $P_{r1}$, which indicates the possible maximum lateral movement of the wheel in relative to the rail. It can be estimated by the maximum instantaneous lateral acceleration $a_{ymax}$ of the wheel in the sample period of the camera. For simplification, $a_{ymax}$ is statistically estimated as a constant value derived from the field measurement data.

**Tab. 3.** Defined rules, indicators and thresholds in the rule engine

| Index | Rules | Indicators | Thresholds |
|---|---|---|---|
| 1 | Y coordinate of the detected points should remain constant in comparison to the reference one (which can be manually defined in the calibration process or using the detection result on the first frame). | RMSE $$\sigma_y = \frac{1}{3}\sqrt{\sum_{i=1}^{3}(y_i - y_{ref})^2}$$ | $\sigma_{yTH_1} = 5\ mm$ $\sigma_{yTH_0} = 10\ mm$ |
| 2 | Y coordinate of the detected points should remain constant in the adjacent frames. | RMSE $$\sigma_{yt} = \frac{1}{3}\sqrt{\sum_{i=1}^{3}(y_{i,t} - y_{i,t-1})^2}$$ | $\sigma_{ytTH_1} = 5\ mm$ $\sigma_{ytTH_0} = 10\ mm$ |
| 3 | The width of the rail head calculated by the x coordinates of the two rail reference points ($P_{r1}$ and $P_{r2}$) should remain constant. | Difference $d_x = |x_{r1} - x_{r2}|$ | $d_{xTH_1} = 5\ mm$ $d_{xTH_0} = 10\ mm$ |
| 4 | The two rail reference points should move in the same lateral direction or remain unchanged in the adjacent frames. | Boolean $(x_{r1,t} - x_{r1,t-1}) \cdot (x_{r2,t} - x_{r2,t-1}) \geq 0$ | $B_{TH_0} = True$ |
| 5 | The wheel lateral displacement between two adjacent frames should be smaller than that calculated by the maximal wheel lateral acceleration. | Lateral displacement $\Delta D = |x_{w,t} - x_{w,t-1}|$ | $\Delta D_{TH_0} = 0.5 \cdot a_{ymax} \cdot \Delta t^2$ |

### 3.4. Image corruption for data augmentation

As the CV system is exposed to a harsh railway environment, a solid housing and an air cleaning system are tailored to protect and clean the camera lenses. Apart from this, we propose a data augmentation procedure during DL model training to enhance the model robustness against possible image corruptions. Taking advantage of previous studies on image corruption [41,42], the relevant corruption types in Fig. 6 are modeled. For a given image $I \in \mathcal{R}^{N \times N}$, $I(x, y)$ in the range $(0, 255)$ denotes the original pixel intensity at the position $(x, y)$. Gaussian noise may arise during optical sensing. The intensity function of the corrupted image $I_{gn}(x, y)$ injected with Gaussian noise is given by

$$I_{gn}(\text{x,y}) = I(x,y)/255 + c \cdot p \tag{4}$$

$$I_{gn}(\text{x,y}) = \begin{cases} 0 & if\ I_{gn}(\text{x,y}) < 0 \\ I_{gn}(\text{x,y}) \cdot 255 & if\ 0 \le I_{gn}(\text{x,y}) \le 255 \\ 255 & if\ I_{gn}(\text{x,y}) > 255 \end{cases} \tag{5}$$

where $c$ is a settable scale representing the severe level and $p$ is the Gaussian distribution.

Shot noise could occur during photon counting in optical systems. The intensity function of the corrupted image $I_{gn}(x, y)$ injected with shot noise is given by

$$I_{sn}(\text{x,y}) = f[c \cdot I(x,y)/255]/c \tag{6}$$

$$I_{sn}(\text{x,y}) = \begin{cases} 0 & if\ I_{sn}(\text{x,y}) < 0 \\ I_{sn}(\text{x,y}) \cdot 255 & if\ 0 \le I_{sn}(\text{x,y}) \le 255 \\ 255 & if\ I_{sn}(\text{x,y}) \end{cases} \tag{7}$$

where $f$ is subject to the Poisson distribution.

The modelled impulsive noise refers to salt-and-pepper noise which could originate from sharp and sudden disturbances in the imaging process. The intensity function of the corrupted image $I_{in}(x, y)$ injected with impulsive noise is given by

$$I_{in}(\text{x,y}) = \begin{cases} 0 & if\ c \cdot \alpha/2 \\ I(x,y) & if\ 1 - c \cdot \alpha \\ 255 & if\ c \cdot \alpha/2 \end{cases} \tag{8}$$

where $\alpha$ is the probability that a pixel is altered.

Defocus blur is that the image is out of focus, which is caused by the fact that the camera integrates the light over areas during sensing. Blur is commonly modelled by convolution of the original image with a uniform point spread function (PSF). The defocus-blurred image $I_{db}$ is given by

$$I_{db} = I * K \tag{9}$$

$$K(x,y) = \begin{cases} 0 & if\ \sqrt{x^2 + y^2} < r \\ 1/\pi r^2 & if\ \sqrt{x^2 + y^2} \ge r \end{cases} \tag{10}$$

where $K$ is the parametric PSF for defocus blur and $r$ is the radius parameter of $K$ and linearly correlated to the severe level $c$.

Motion blur occurs when the vehicle is excited by large track/rail irregularities. The linear-motion-blurred image $I_{mb}$ is given by

$$I_{mb} = I * K \tag{11}$$

$$K(x,y) = \begin{cases} 1/r & if\ 0 \le x \le r \\ 0 & otherwise \end{cases} \tag{12}$$

where $K$ is the parametric PSF for linear motion blur and $r$ denotes the extent of the motion blur, relying on the severe level $c$.

In addition, several weather conditions are modeled. Snowy scenes are generated by randomly adding white motion-blurred particles and whitening the entire image. The image with frost is an overlay of the original image and several template images of frosted glass. Fog is modeled by plasma fractal using the diamond-square algorithm. Sunny/shady effect is simulated by increasing/decreasing the brightness of the original image, where the pixel intensity of the first channel in the HLS color space of the image is altered. Furthermore, several common augmentation techniques are applied, such as horizontal flip, rotation and occlusion. In addition, we mimic the images taken at different camera positions and orientations. For each original 1920×1080 image, we randomly crop the 256×256 RoI at different positions. Afterward, point perspective transformation is applied to simulate the variations of the camera's orientation.

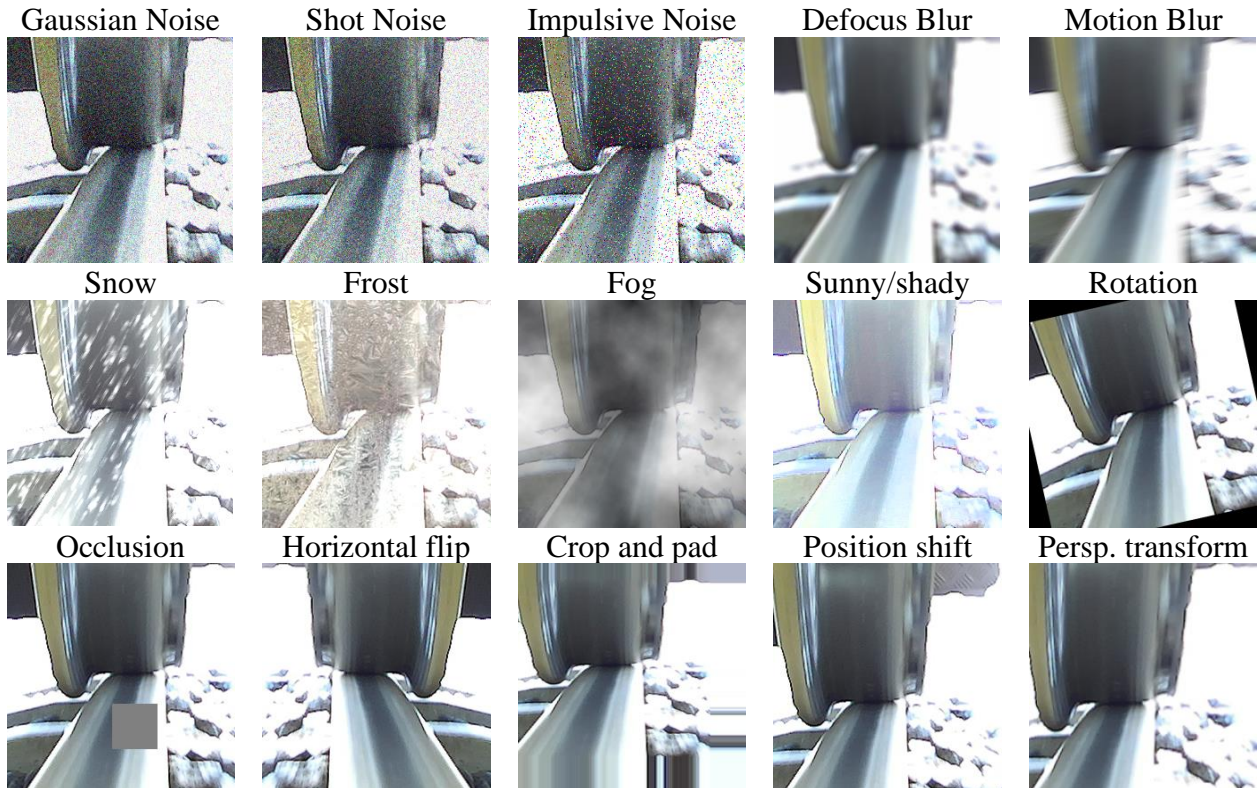

**Fig. 6.** Synthetic images for data augmentation

## 4. Experiments and results

### 4.1. Field tests and datasets

We conducted three field tests under different operational conditions in Italy and Lithuania. The first two tests were performed for data collection and algorithm development at low running speeds. Afterward, the last test was carried out as the validation test on both straight and curved track sections at regular operating speeds (up to 100 km/h). In Italy, the prototype of the CV system has been installed on the bogie frame of Aldebaran 2.0, which is the newest track recording coach of Rete Ferroviaria Italiana (RFI, i.e. Italian infrastructure manager) equipped with a commercial TG measurement system, as shown in Fig. 7. (a). The first test consisted of several runs within the workshop plant in Catanzaro Lido on both straight and curved track sections. The curved track sections correspond to two switches with a curve radius of 170 m and a tangent of 0.12. During the field test, the Aldebaran 2.0 coach was driven by a locomotive at low speeds (between 2 and 10 km/h). We test different conditions, i.e. two lateral positions of the camera with respect to the wheel and four camera configurations for different resolutions and sample rates. The video data from 3 test runs are used for model training, while 3 test runs are used for testing.

In Lithuania, the second test was performed on the mainline in the vicinity of Vilnius. Two CV systems were installed on the bogie frame of a track recording coach operated by Lithuanian Railways, see Fig. 7. (b). The videos for both wheels were recorded simultaneously. Two forward runs at speeds of around 20 km/h and one backwards run at lower speeds were conducted. The camera setting remains unchanged during the test runs. One forward run is used for training, while the other data is used for testing.

The last validation test was performed in Italy on the same vehicle type of Aldebaran 2.0. The vehicle has been operated in regular operating conditions up to 100 km/h between Rome and Pisa for two days.

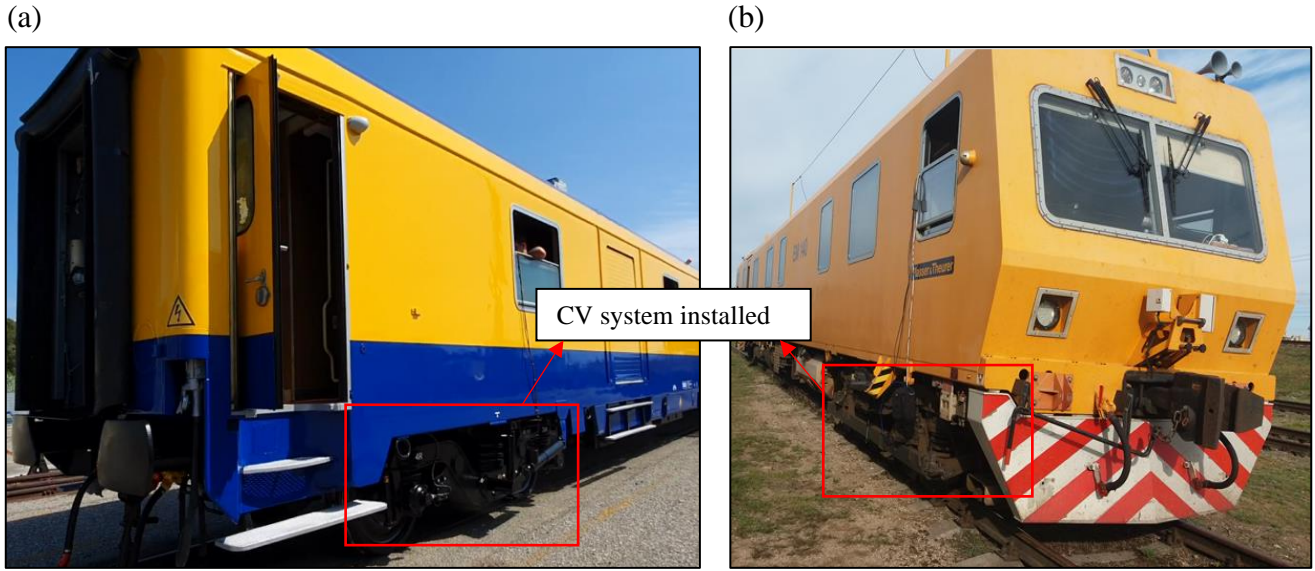

**Fig. 7.** (a) CV system installed on a track recording coach in Italy; (b) CV system installed on a track recording coach in Lithuania

The algorithm development is based on the data obtained in the first two tests. As the video data was sampled at 30 fps and the vehicle ran at low speeds, there are a large number of duplicate frames in the video. To build the dataset, we select one image per 30 frames from the video data collected in Lithuania, while one image per 60 frames from the video data collected in Italy. Other images originate from static tests at other locations and a relevant Youtube video [43]. In static tests, the same ZED2 stereo camera was used for image capture. The images of different bogies standing on the track were obtained, examples of which are shown in Fig. 14 in Annex I. The Youtube video was filmed by a GoPro camera during a regular railway operation. The video frames were extracted as shown in Fig. 15 in Annex I. The defined virtual points were manually annotated on the original images of 1920×1080 pixels. The coordinates of the labeled points are the ground truth for CNN training. We have 767 annotated images in total. In order to increase the amount of the annotated data, we generate five 256×256 images of RoI cropped at different positions on each original image. In this way, we have 3835 labeled images. They are split into a training dataset, a validation dataset and a test dataset with the ratio of 6/2/2, namely 2301 images for training and 767 images for validation and testing respectively. We conduct extensive experiments to validate the proposed approach as follows.

### 4.2. Training and evaluation of YOLOv3-tiny for calibration

In YOLOv3-tiny, we merely modify the YOLO layers for RoI detection, while the first 13 layers, i.e. Darknet, have not been changed. This allows us to transfer the pre-trained weights of Darknet to the modified YOLOv3-tiny. In this way, the model for RoI detection can be trained with fewer annotated images. Fig. 8 presents the pipeline for training and evaluation of YOLOv3-tiny on our datasets. YOLOv3-tiny is first pretrained on the COCO dataset [44], which contains 123287 annotated images in total, incl. 3745 images related to the railway. The learned parameters of Darknet are transferable, while the learned parameters of the YOLO layers are discarded. Our training dataset consists of 800 images from static tests and the Youtube video. The raw 1920×1080 images are resized to the 416×416 ones, fed into YOLOv3-tiny. The pre-trained YOLOv3-tiny is trained with adaptive moment estimation (Adam) for 30 epochs, which is a gradient-descent-based optimization algorithm. Afterward, the trained model of YOLOv3-tiny is evaluated on 767 annotated images for keypoint detection. The evaluation metric is whether the labeled keypoints are inside the predicted bounding box within an image. YOLOv3-tiny has achieved a detection accuracy of 100%.

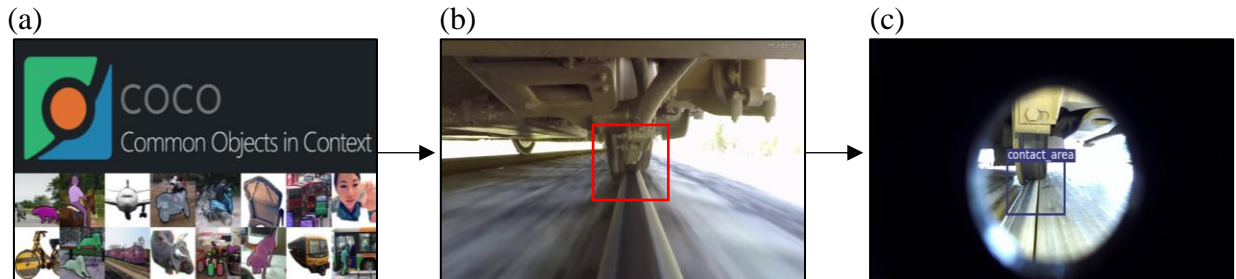

**Fig. 8.** Pipeline for training and evaluation of YOLOv3-tiny: (a) Pre-training on COCO dataset; (b) Fine-tuning on our training dataset; (c) Evaluation on our test dataset

### 4.3. Training and evaluation of LightPointNet

The evaluation of LightPointNet is conducted in a two-fold way. Firstly, LightPointNet is trained and evaluated on the individual images randomly selected and cropped from the video frames. The evaluation metric is the deviation in terms of pixels between the ground-truth and the predicted x-coordinate of $P_w$ and $P_{r1}$. We compare the evaluation results of LightPointNet with those of the baselines. Secondly, the trained LightPointNet is applied on the video sequences. The evaluation metric is defined as the count that detects the predictions exceeding the thresholds $TH_0$ in the rule engine.

For comparison, we implement three DL-based baselines, i.e. PoseResNet [28], ReceptionNet [30] and BlazePose [31], as well as a method using conventional CV techniques developed in our previous work [32]. PoseResNet is the representative of the detection-based HPE network. ReceptionNet is similar to our solution, using softmax for coordinate regression. The main difference lies in the encoder and decoder for feature extraction. BlazePose is a hybrid solution, requiring two-step training. In the first step, the network is trained on the heatmap branch as the common detection-based HPE networks do. In the second step, the weights in the heatmap branch are frozen, while the regression branch is activated for training. In the inference stage, the regression branch is applied to directly output the point coordinates, which avoids the quantization errors and reduces the computation complexity. Our previous method mainly uses template matching to track the wheel flange and line tracking to track the rail line. More details can be found in [32]. However, this method is only applicable to track the points in the video sequences by manually selecting the reference template and line on the first frame. It is not able to automatically detect the wheel and rail.

In the first evaluation experiment, LightPointNet is compared with PoseResNet, ReceptionNet and BlazePose. The DL networks are trained from scratch on our training dataset (incl. 2301 256×256 images) and evaluated on the testing dataset (incl. 767 images). The validation dataset is used to prevent overfitting by evaluating the temporary model trained in each epoch during the training process. Adam with the multistage learning rate is applied to minimize the integral loss over 100 epochs. We repeat the training process using different random seeds five times and select the best models for the comparison. The main reason is that CNN learnable weights are randomized at initialization and learning samples are randomized. We get models that perform differently with the same training conditions. The details of implementation and experiment settings can be found in our code repository.

(a)

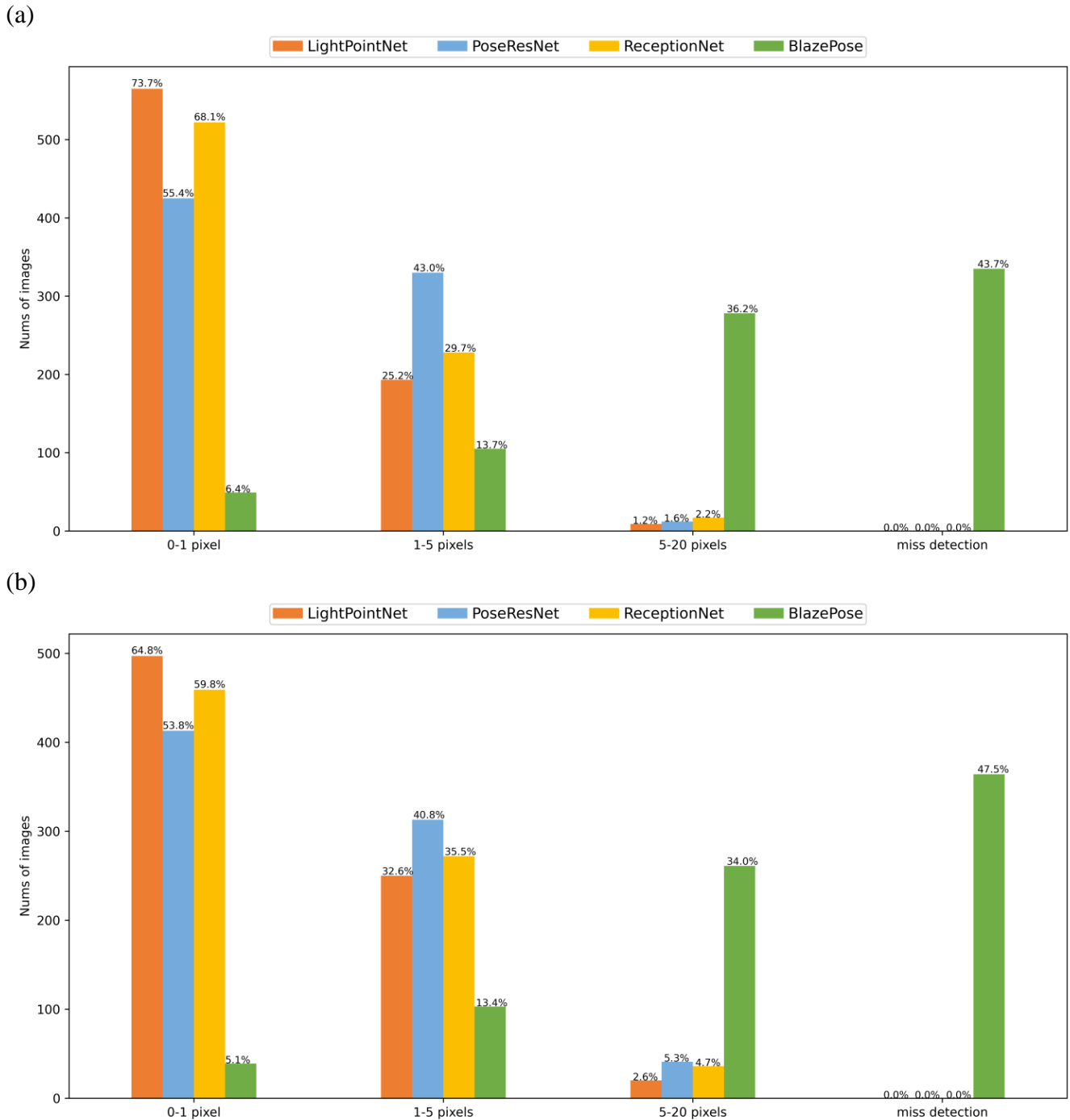

(b)

**Fig. 9.** (a) detection errors (in pixels) for the wheel reference point; (b) detection errors (in pixels) for the rail reference point

The evaluation metric is the deviation in terms of pixels between the ground-truth and the predicted x-coordinate of $P_w$ and $P_{r1}$. We divide detection errors into four groups. "0-1 pixel" means either no error or an error of 1 pixel. In our case, 1 pixel means 0.78 mm (depending on the image resolution as well as the distance between the camera and the wheel). A small error of 1-5 pixels is tolerable. A large error of 5-20 pixels is unacceptable. An error with more than 20 pixels is defined as "miss detection". Fig. 9 shows the detection errors for the wheel and rail reference points. Overall, LightPointNet achieves the best performance among the baselines. The error rate for the wheel reference point is lower than that for the rail reference point. Comparing LightPointNet with the baselines, we observe at first glance that BlazePose fails to accurately detect the wheel and rail reference point using the regression branch. According to our experiment results, the heatmap branch of BlazePose can achieve comparable performance as PoseResNet does. Switching the heatmap branch to the regression branch induces significant performance loss. Second, LightPointNet and ReceptionNet, which involve integral regression from heatmaps, deliver fewer detection errors than PoseResNet, which directly outputs heatmaps and thus induces the quantization errors. Third, LightPointNet overperforms ReceptionNet. The reason may lie in the different architectures of the encoder and decoder.

Furthermore, we evaluate LightPointNet and our previous method [32] on video sequences. As the ground-truth points have not been manually labeled on video sequences, a rigorous validation comparing the prediction with the ground truth cannot be performed. In this case, the rule engine is used for the evaluation. The evaluation metric is defined as the count that detects the predictions exceeding the thresholds in the rule engine. We select several video sequences representing different track layouts. Tab. 4 compares the results of two methods, where the baseline refers to our previous method. The baseline and LightPointNet can achieve comparable results on the videos obtained in Lithuania. However, LightPointNet significantly overperforms the baseline in the first test in Italy. In these videos, brightness is much lower and the backgrounds are more complicated than those in Lithuania. The baseline almost fails to track the rail lines in the videos and thus fails to deliver reliable results, although much effort has been paid to tune the relevant parameters for pre-processing. In the last validation test, a part of the video sequences, named "Italy2_highspeed", has been randomly selected, where the vehicle has reached the maximum speed of 100 km/h. The illumination has been improved during this test. LightPointNet delivers 3.51% miss detection rate, while the baseline has 18.49%, which is much better than that in the first test. A high speed results in motion blur of the complex backgrounds, which makes it easier to track the wheel temple and rail line. It is worth noting that the current generalization ability of the LightPointNet model trained on the aforementioned training dataset was not sufficient to achieve a high detection accuracy on a new test dataset. We randomly labeled 351 images in the validation test for model fine-tuning. To prevent data leakage, we avoided using any images from the test video sequence "Italy2_highspeed". The number of the labeled images is merely 0.1% of the total frames obtained in this validation test.

**Tab. 4.** Test results of LightPointNet and the baseline on the representative video sequences

| Video sequence | Frame number | Model | Implausible detection exceeding $TH_0$ | Percentage of Implausible detection |
|---|---|---|---|---|
| Lithuania_forward _straight | 4997 | LightPointNet | 207 | **4.45%** |
| | | Baseline | 245 | 4.90% |
| Lithuania_backward _straight | 7996 | LightPointNet | 84 | 1.05% |
| | | Baseline | 78 | **0.98%** |
| Italy_curve | 2586 | LightPointNet | 200 | **7.73%** |
| | | Baseline | 2263 | 87.5% |
| Italy_swtich | 492 | LightPointNet | 50 | **10.16%** |
| | | Baseline | 344 | 69.92% |
| Italy_straight | 4024 | LightPointNet | 233 | **5.79%** |
| | | Baseline | 3144 | 78.13% |
| Italy2_highspeed | 14082 | LightPointNet | 495 | **3.51%** |
| | | Baseline | 2604 | 18.49% |

*4.4. Evaluation of rule engine*

In order to evaluate the effectiveness of the proposed rule engine, we evaluate our approach with and without the rule engine on the video sequences. In the approach without the rule engine, LightPointNet is applied for each video frame and directly outputs the predicted coordinates as the final results. The percentage of implausible detection has been shown in Tab. 4. In the approach with the rule engine, LightPointNet's outputs are fed into the rule engine. The corner cases are detected and the corresponding implausible results are discarded or corrected. Fig. 10 illustrates several typical corner cases where LightPointNet fails to deliver a reliable detection result. The corner cases may have specific scenes in backgrounds like workshops and platforms. The switch and crossing zones have a unique track layout that may mislead the detector. A wheel bounce results in a sudden change of the y-coordinate of $P_w$ and trigger the rule engine. It may also cause miss detection of LightPointNet due to the strong motion blur. Such corner cases will be added to the training dataset for further model training.

For evaluation of the correction mechanism, the trajectory of the LDWR over the frames is displayed. Fig. 11 shows the trajectory with and without the rule engine calculated on the video sequence "Italy_straight" containing 4024 frames. The correction mechanism based on the rule engine uses the information of two adjacent frames to remove the coordinates' unreliable sudden changes, as shown by red impulses in Fig. 11. Nevertheless, tracking the actual lateral movement of the wheel has not been affected. For instance, a sizeable lateral wheel movement occurs between the 2750th and 2950th frame is visible by the blue line in Fig. 11. However, we observe that the predicted coordinates' small-scale turbulence cannot be smoothed by the rule engine. The data fusion with the corresponding wheel accelerations may cover this gap.

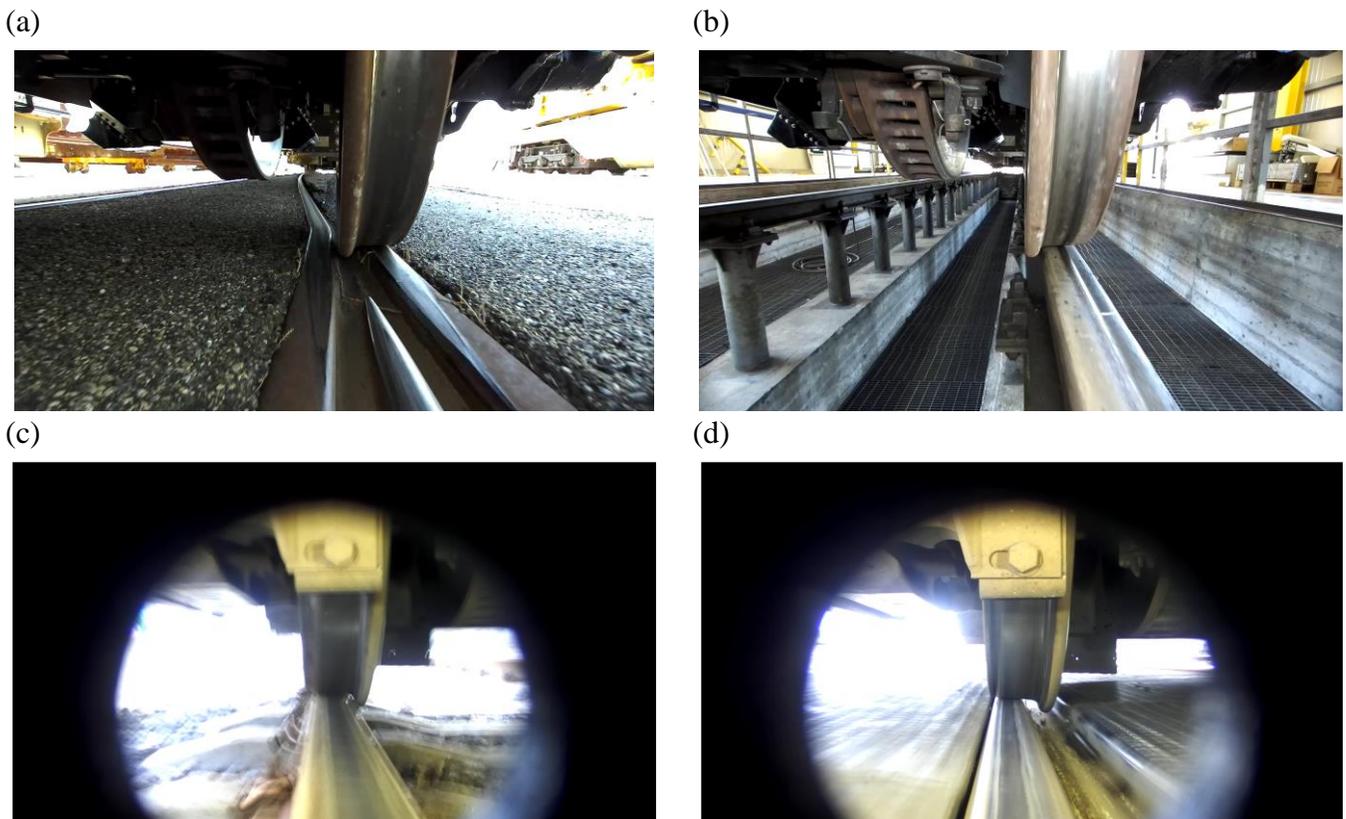

**Fig. 10.** Corner cases detected by the rule engine: (a) switch & crossing zone; (b) workshop zone; (c) wheel bounce; (d) platform zone

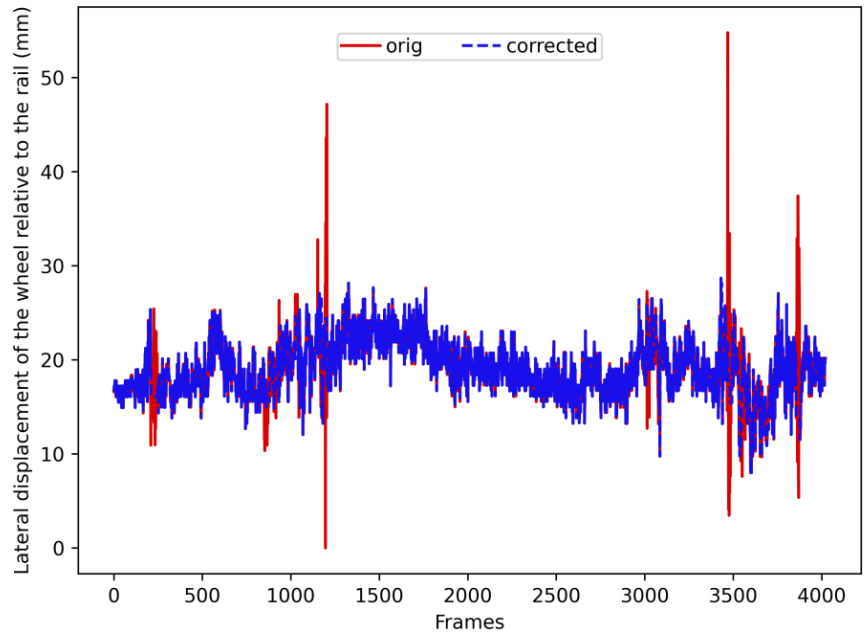

**Fig. 11.** Comparison of the measured lateral displacement of the wheel relative to the rail with (blue) and without (red) the rule engine based correction mechanism

### 4.5. Evaluation of the entire approach

The entire algorithm is executed on the Nvidia Jetson TX2 platform in real-time. The tracking results on two video sequences are recorded as the demonstration videos. As the points are hardly visible on the raw 1920×1080 images, the demo videos merely display the 256×256 RoI, which is automatically detected by YOLOv3-tiny at the first step of the proposed approach. Fig. 12 shows the tracked points $P_w$ and $P_{r1}$ on the wheel flange and the rail edge in the RoI for the three field tests. These two points are used for the calculation of the lateral wheel displacement. $P_{r2}$ is on the other side of the rail edge and provides the geometric information for the rule engine. It is not displayed on the demo videos.

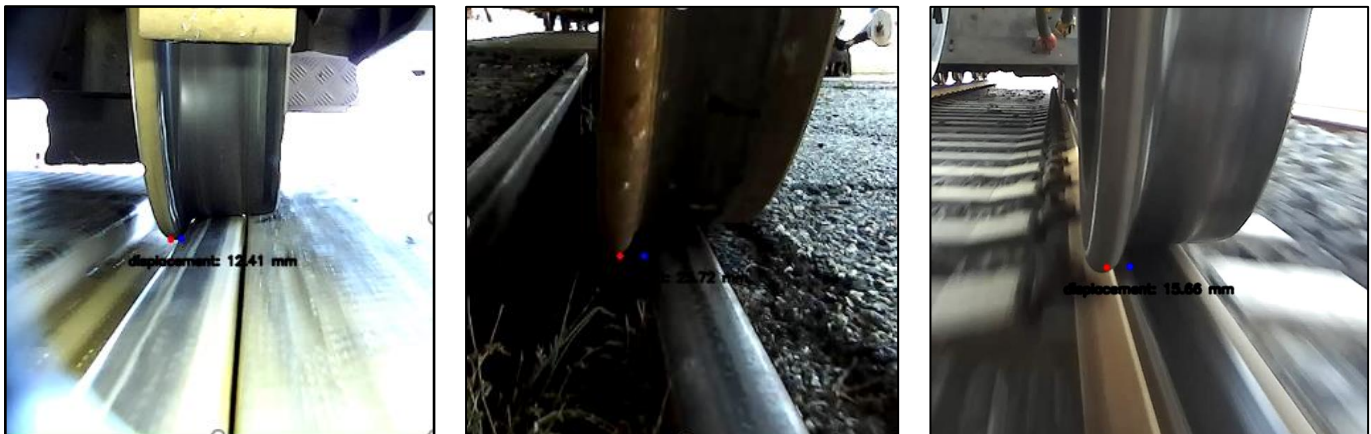

**Fig. 12.** Real-time lateral displacement measurement of the wheel on the rail to support track geometry monitoring (screenshot from the demo videos *https://youtu.be/it21cE87LCM*, *https://youtu.be/Nc1bkQdkkSM* and https://youtu.be/K5HnUixyGxo

Wheel's lateral motion has been successfully tracked by tracking the virtual points. However, we observe slight shifts of the detected virtual points in the lateral direction, although the wheel's actual position does not change. It results in sudden changes of lateral displacement in several millimeters. This indicates a measurement uncertainty up to 2 mm based on our observation, which stems from point tracking and displacement calculation. In our previous study [32], the displacement calculation method based on two reference points have been tested in a laboratory, where the stereo camera was placed at different distances and view angles with respect to a standard gauge block. Two reference points were manually selected on the gauge block's left and right edge to calculate its width. The measurement uncertainty (i.e. in the form of standard deviation) has been determined as 0.4 mm. Therefore, we conclude that the point detection of LightPointNet induces the main uncertainty. On the one hand, this is due to the model's performance limitation trained on the currently collected training data. On the other hand, the uncertainty could originate from label noise, which occurs when we manually annotate the virtual points as the ground truth. Due to the complex background, variable illumination conditions, and labeling tool restrictions, the accurate point position on the wheel flange and the railhead edge can hardly be determined. An annotation deviation of several pixels on a similar video frame is quite common. For the 1920×1080 resolution and the distance between the camera and the wheel, one pixel refers to 0.78 mm. Therefore, a measurement uncertainty of 2 mm due to manual annotation is understandable and can be hardly avoided. A possible solution is to increase the image resolution of RoI. In future work, we consider replacing the current camera with the one having a narrower field of view and closer focusing distance.

### 4.6. Computational complexity and real-time capability

CNN's computational complexity can be theoretically estimated by the number of parameters and floating-point operations (FLOPs). A regular convolution layer consists of $N$ convolutional filters, each of which is parameterized by the weights $W \in \mathcal{R}^{K \times K}$, where $K$ denotes the width. When it takes a feature map $F_{in} \in \mathcal{R}^{D \times D \times C}$ as the input and outputs a feature map $F_{out} \in \mathcal{R}^{D \times D \times N}$, the total parameters $P_c$ and FLOPs are given by formulas (14) and (15), where the parameter number of bias and the accumulated operation are neglected.

$$P_c = K \times K \times C \times N \tag{13}$$

$$FLOPs_c = 2MAC = 2 \times (K \times K \times C \times N \times D \times D) \tag{14}$$

In LightPointNet, regular $K \times K$ convolution is replaced with the combination of $1 \times 1$ pointwise convolution and $K \times K$ depthwise convolution, which is named as the depthwise separable convolution (DSC). Its parameter numbers $P_{DSC}$ and FLOP$_{DSC}$ are significantly reduced, given by:

$$P_{DSC} = K \times K \times C \tag{15}$$

$$FLOPs_{DSC} = 2 \times (K \times K \times C \times D \times D + N \times C \times D \times D) \tag{16}$$

The reduction ratio in parameter $r_P$ and in operation $r_{FLOPs}$ are given by:

$$r_P = \frac{1}{N} \tag{17}$$

$$r_{FLOPs} = \frac{1}{N} + \frac{1}{K^2} \tag{18}$$

The computational complexity of a conventional image processing algorithm can be hardly accurately measured. In the baseline, we mainly use a template matching algorithm for wheel tracking and a line tracking algorithm provided in the Visual Servoing Platform library for rail tracking. Its theoretical computational complexity can hardly be calculated. For a more accurate comparison, the actual time consumption, i.e. latency, is measured for each algorithm. The latency relies on the hardware and software platform. In our application, we implement the DL models in PyTorch 1.9 (which is an open-source DL framework) and deploy the models on the edge computer Nvidia Jetson TX2 for inference. The baseline is implemented with OpenCV libraries. We measure the time consumption on this platform and calculate frame per second (FPS) averaged over the testing video sequences as the evaluation metric. This allows the comparison between DL models and the baseline.

LightPointNet uses several lightweight measures to reduce the number of parameters and FLOPs while maintaining network performance, incl. using filters of small sizes, using DSC, using SE modules and linear bottleneck structure. More details of the lightweight measures can be found in our previous study [46]. We compare our LightPointNet with the baselines to show the effectiveness of lightweight. Tab. 5 shows the computational complexity of different models with a batch size of 16. Parameters and FLOPs of DL models are measured by a third-party tool. The third row indicates the latency in fps of the original DL models implemented in Pytorch. The fourth row indicates the latency of the DL models in the format of Nvidia TensorRT, which will be explained later. At first glance, we find that the baseline has a larger latency than our LightPointNet in terms of FPS. However, both are slower than the real-time requirement (i.e. 30 FPS). Comparing LightPointNet with PoseResnet and BlazePose, the latency of LightPointNet is slightly less than that of PoseResnet and BlazePose, although FLOPs of LightPointNet are much lower. It indicates that the platform-dependent latency is also much affected by other factors apart from FLOPs. In terms of parameters, LightPointNet has almost 12-times fewer parameters than PoseResnet, which means much less memory usage. A Pytorch model of LightPointNet occupies 11 MB, whereas a PyTorch model of PoseResnet occupies 130 MB.

**Tab. 5.** Computational complexity of LightPointNet and the baselines (M for million, G for Giga)

| Results | LightPointNet | PoseResnet | ReceptionNet | BlazePose | Baseline |
|---|---|---|---|---|---|
| **Parameters** | 2.86 M | 33.99 M | 5.97 M | 34.00 M | - |
| **FLOPs** | 5.51 G | 12.90 G | 6.67 G | 17.39 G | - |
| **FPS** | 20 | 16 | 19 | 15 | 18 |
| **FPS (TensorRT)** | 39 | 26 | 35 | 23 | - |

We observe that none of the PyTorch models has a real-time ability on the target platform. To further reduce the latency, we transform the PyTorch models into the format of TensorRT, which speeds up the inference of a DL model on Nvidia's GPUs. TensorRT forces the models for low precision inference. The learned parameters of weights and biases within a NN are typically represented in the format of float32, occupying 32 bits. TensorRT transforms these parameters into the 8-bit representation. This dramatically accelerates the inference process by sacrificing little accuracy. Furthermore, TensorRT optimizes the computation graph of a NN to accelerate the computation. More details can be found in [47]. The last row of Tab. 5 shows the latency of the DL models in the TensorRT format. LightPointNet and ReceptionNet can satisfy the real-time requirement.

### 4.7. Data augmentation for robustness enhancement

The model robustness plays an essential role in harsh outdoor conditions. The degradation or interference of sensors may result in image noise. Large vibrations induced by severe track/rail irregularities may result in image blur. Dirt and dust on camera lenses may result in occlusions in images. Varying weather conditions may result in variations of intensity distributions within images. Based on these types of image corruption, we build a corrupted testing dataset. Each image from the original test dataset containing 767 images is augmented with a corruption method randomly selected from the ones shown in Fig. 6. Each corruption method contains a severity scale $c$, which controls the severity of the corruption. The scale $c$ is randomly set in the range from 1 to 3. First, we investigate the model robustness against image corruption. LightPointNet and the baselines are trained on the clean training dataset without corrupted images and tested on the corrupted test dataset. Comparing Fig. 13 (a) with Fig. 9, we observe that the model performance of LightPointNet dramatically drops from around 70% to 23.2% of "0-1 pixel" detection errors. ReceptionNet achieves similar performance as LightPointNet. This indicates the insufficient robustness of all the DL models against the potential image corruptions in harsh outdoor conditions. Afterward, we investigate the effect of data augmentation by repeating the corruption process on the training dataset as data augmentation.

The DL models are trained on the augmented training dataset and tested on the corrupted test dataset. Fig. 13 (b) shows data augmentation largely reduces the miss detection rate of LightPointNet, PoseResNet and ReceptionNet. In particular for LightPointNet, the rate of the "0-1 pixel" errors is improved by 6.5%. On the one hand, this result proves the effect of data augmentation and superiority of LightPointNet. On the other hand, data augmentation alone is not sufficient to ensure robustness, since the improvement is merely a drop in the bucket in comparison to performance loss induced by image corruption. Other generalization measures such as domain generalization should be combined.

(a)

(b)

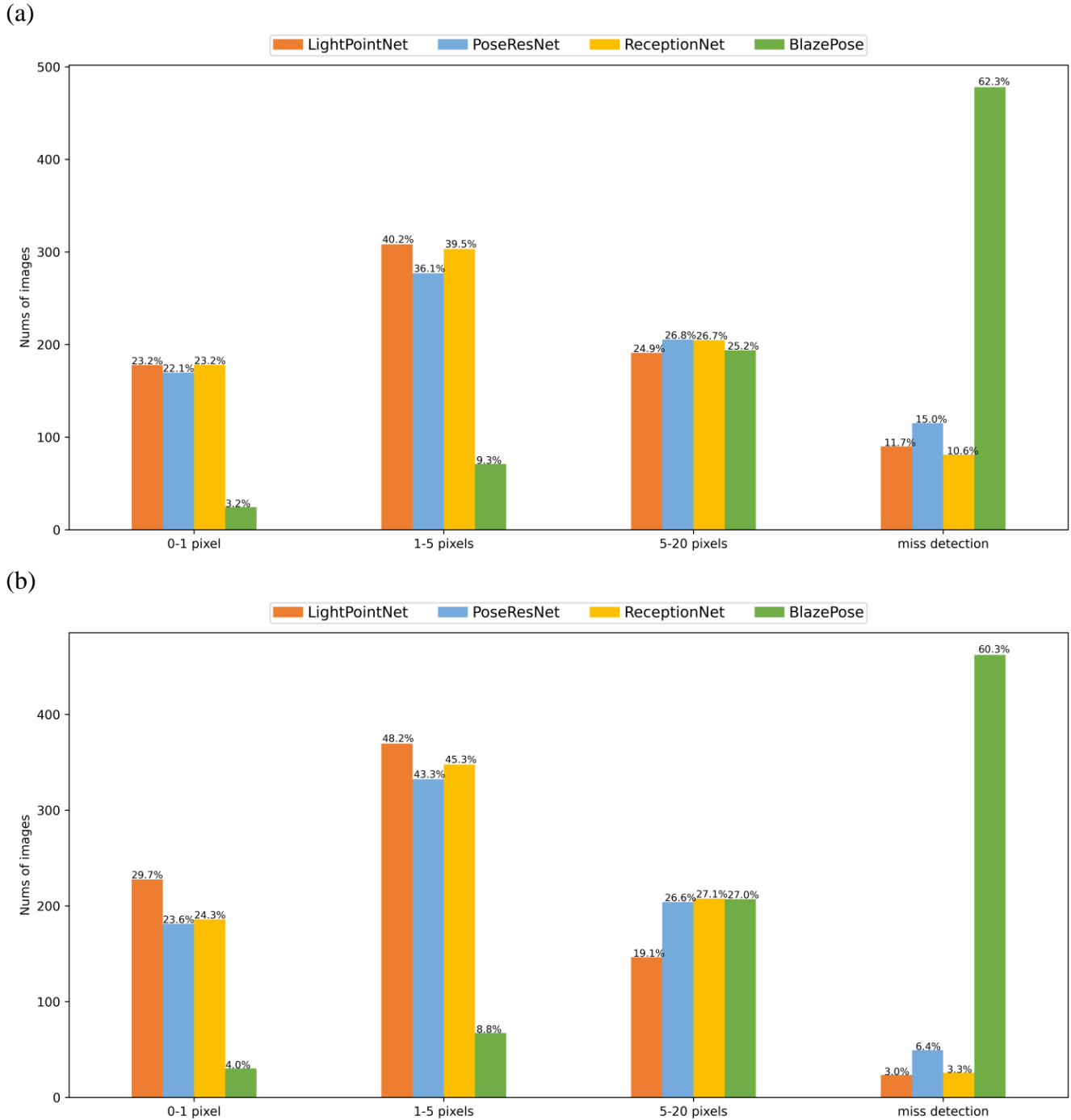

**Fig. 13.** detection errors (in pixels) averaged over three reference points (a) when the models are trained on the cleaning training dataset and tested on the corrupted test dataset; (b) when the models are trained on the augmented training dataset and tested on the corrupted test dataset

## 5. Conclusions

The virtual point tracking approach was proposed to tackle the issue of dynamic displacement measurement with varying and noisy backgrounds. The entire approach has been validated and demonstrated for lateral displacement measurement of the wheelsets on the rail tracks, in order to support track geometry monitoring on in-service rail vehicles. The feasibility of the proposed solution has been demonstrated in the field tests under regular railway operating conditions. It can satisfy the real-time processing requirement and achieve a measurement uncertainty of up to 2 mm. The core component of our approach is LightPointNet for point detection, which is a lightweight CNN architecture. It outperforms several baselines using either conventional image processing techniques or other deep learning networks. One unsolved issue in this work is the robustness against harsh outdoor conditions and the generalization ability, which are the common issue for arbitrary machine learning methods. They will be addressed in our future work.

### Declaration of Competing Interest

On behalf of all authors, the corresponding authors state that there is no conflict of interest.

### Acknowledgments

The research is funded by the EU Shift2Rail project Assets4Rail (Grand number: 826250) under Horizon 2020 Framework Programme.

### References

[1] Y. Xu, J.M.W. Brownjohn, Review of machine-vision based methodologies for displacement measurement in civil structures, Journal of Civil Structural Health Monitoring. 8 (2017) 91–110. doi:10.1007/s13349-017-0261-4.

[2] D. Zhan, D. Jing, M. Wu, D. Zhang, L. Yu, T. Chen, An Accurate and Efficient Vision Measurement Approach for Railway Catenary Geometry Parameters, IEEE Transactions on Instrumentation and Measurement. 67 (2018) 2841–2853. doi:10.1109/tim.2018.2830862.

[3] Z.-W. Li, Y.-L. He, X.-Z. Liu, Y.-L. Zhou, Long-Term Monitoring for Track Slab in High-Speed Rail via Vision Sensing, IEEE Access. 8 (2020) 156043–156052. doi:10.1109/access.2020.3017125.

[4] J. Baqersad, P. Poozesh, C. Niezrecki, P. Avitabile, Photogrammetry and optical methods in structural dynamics – A review, Mechanical Systems and Signal Processing. 86 (2017) 17–34. doi:10.1016/j.ymssp.2016.02.011.

[5] C.-Z. Dong, F.N. Catbas, A review of computer vision–based structural health monitoring at local and global levels, Structural Health Monitoring 20 (2021) 692–743. https://doi.org/10.1177/1475921720935585.

[6] S. Wang, H. Wang, Y. Zhou, J. Liu, P. Dai, X. Du, M. Abdel Wahab, Automatic laser profile recognition and fast tracking for structured light measurement using deep learning and template matching, Measurement 169 (2021) 108362. https://doi.org/10.1016/j.measurement.2020.108362.

[7] T. Jiang, G.T. Frøseth, A. Rønnquist, E. Fagerholt, A robust line-tracking photogrammetry method for uplift measurements of railway catenary systems in noisy backgrounds, Mechanical Systems and Signal Processing. 144 (2020) 106888. doi:10.1016/j.ymssp.2020.106888.

[8] European Committee for standardization. EN 13848-1: Railway applications - Track - Track geometry quality - Part 1: Characterization of track geometry. (2019)

[9] V. Skrickij, D. Shi, M. Palinko, L. Rizzetto, G. Bureika, Wheel-rail transversal position monitoring technologies. Technical Report Deliverable D8.1, EU Horizon 2020 project Assets4Rail. (2019). http://www.assets4rail.eu/results-publications/. (accessed December 7, 2020).

[10] B. Ripke et al., Report on track/switch parameters and problem zones, technical report Deliverable D4.1 of the IN2SMART project, https://projects.shift2rail.org/s2r_ip3_n.aspx?p=IN2SMART (accessed December 3, 2020).

[11] I. Soleimanmeigouni, A. Ahmadi, H. Khajehei, A. Nissen, Investigation of the effect of the inspection intervals on the track geometry condition. Structure and Infrastructure Engineering. 16 (2020) 1138–1146. doi: 10.1080/15732479.2019.1687528.

[12] P. Weston, C. Roberts, G. Yeo, E. Stewart, Perspectives on railway track geometry condition monitoring from in-service railway vehicles. Vehicle System Dynamics. 53(2015) 1063-1091. doi:10.1080/00423114.2015.1034730

[13] H. True, L.E. Christiansen, Why is it so difficult to determine the lateral Position of the Rails by a Measurement of the Motion of an Axle on a moving Vehicle? Proceedings of First International Conference on Rail Transportation. (2017)

[14] A.D. Rosa, S. Alfi, S. Bruni, Estimation of lateral and cross alignment in a railway track based on vehicle dynamics measurements, Mechanical Systems and Signal Processing. 116 (2019) 606–623. doi:10.1016/j.ymssp.2018.06.041.

[15] A.D. Rosa, R. Kulkarni, A. Qazizadeh, M. Berg, E.D. Gialleonardo, A. Facchinetti, et al., monitoring of lateral and cross level track geometry irregularities through onboard vehicle dynamics measurements using machine learning classification

algorithms, Proceedings of the Institution of Mechanical Engineers, Part F: Journal of Rail and Rapid Transit. (2020) 095440972090664. doi:10.1177/0954409720906649.

[16] J. Sun, E. Meli, W. Cai, H. Gao, M. Chi, A. Rindi, S, Liang, A signal analysis based hunting instability detection methodology for high-speed railway vehicles, Vehicle System Dynamics. (2020). doi: 10.1080/00423114.2020.1763407

[17] M. Kim, Measurement of the Wheel-rail Relative Displacement using the Image Processing Algorithm for the Active Steering Wheelsets, International Journal of Systems Applications, Engineering & Development 6 (2012)

[18] SET Limited, Laser triangulation sensors measure lateral position of rail bogie wheels, Laser Triangulation Sensors Measure Lateral Position of Rail Bogie Wheels, Engineer Live. https://www.engineerlive.com/content/laser-triangulation-sensors-measure-lateral-position-rail-bogie-wheels (accessed December 5, 2020).

[19] D. Yamamoto, Improvement of method for locating position of wheel/rail contact by means of thermal imaging, Quarterly Report of RTRI (2019)

[20] J. Guo, C. Zhu, Dynamic displacement measurement of large-scale structures based on the Lucas–Kanade template tracking algorithm, Mechanical Systems and Signal Processing. 66-67 (2016) 425–436. doi:10.1016/j.ymssp.2015.06.004.

[21] Y.J. Cha, J.G. Chen, O. Büyüköztürk, Output-only computer vision based damage detection using phase-based optical flow and unscented Kalman filters, Engineering Structures. 132(2017) 300-313. doi: https://doi.org/10.1016/j.engstruct.2016.11.038

[22] C.-Z. Dong, O. Celik, F.N. Catbas, E. OBrien, S. Taylor, A Robust Vision-Based Method for Displacement Measurement under Adverse Environmental Factors Using Spatio-Temporal Context Learning and Taylor Approximation, Sensors (Basel) 19 (2019). https://doi.org/10.3390/s19143197.

[23] R. Yang, S.K. Singh, M. Tavakkoli, N. Amiri, Y. Yang, M.A. Karami, et al., CNN-LSTM deep learning architecture for computer vision-based modal frequency detection, Mechanical Systems and Signal Processing. 144 (2020) 106885. doi:10.1016/j.ymssp.2020.106885.

[24] J. Liu, X. Yang, Learning to See the Vibration: A Neural Network for Vibration Frequency Prediction, Sensors. 18 (2018) 2530. doi:10.3390/s18082530.

[25] Y. Chen, Y. Tian, M. He, Monocular human pose estimation: A survey of deep learning-based methods, Computer Vision and Image Understanding. (2020). https://www.sciencedirect.com/science/article/abs/pii/S1077314219301778 (accessed December 7, 2020).

[26] Microsoft, Azure Kinect body tracking joints, Microsoft Docs. (2019). https://docs.microsoft.com/en-us/azure/kinect-dk/body-joints (accessed December 7, 2020).

[27] A. Newell, K. Yang, J. Deng, Stacked hourglass networks for human pose estimation, European conference on computer vision. (2016) 483–499.

[28] B. Xiao, H. Wu, Y. Wei, Simple Baselines for Human Pose Estimation and Tracking, 2018 IEEE/CVF Conference on Computer Vision and Pattern Recognition (2018) 472-487.

[29] K. Sun, B. Xiao, D. Liu, J. Wang, Deep High-Resolution Representation Learning for Human Pose Estimation, 2019 IEEE/CVF Conference on Computer Vision and Pattern Recognition (CVPR) (2019): 5686-5696.

[30] D. Luvizon, D. Picard, H. Tabia, 2D/3D Pose Estimation and Action Recognition using Multitask Deep Learning, 2018 IEEE/CVF Conference on Computer Vision and Pattern Recognition (2018) 5137-5146.

[31] V. Bazarevsky, I. Grishchenko, K. Raveendran, T. Zhu, F. Zhang, M. Grundmann, BlazePose: On-device Real-time Body Pose tracking. arXiv.org. (2020). https://arxiv.org/abs/2006.10204. (accessed December 7, 2020).

[32] V. Skrickij, E. Šabanovič, D. Shi, S. Ricci, L. Rizzetto, G. Bureika, Visual Measurement System for Wheel-Rail Lateral Position Evaluation, Sensors (Basel) 21 (2021). https://doi.org/10.3390/s21041297.

[33] Stereolabs, Datasheet ZED2 Nov 2019 rev6 - Stereolabs, (2019). https://www.stereolabs.com/assets/datasheets/zed2-camera-datasheet.pdf (accessed December 9, 2020).

[34] NVIDIA, NVIDIA Jetson TX2: High Performance AI at the Edge, NVIDIA. (n.d.). https://www.nvidia.com/en-us/autonomous-machines/embedded-systems/jetson-tx2/ (accessed December 9, 2020).

[35] J. Redmon, A. Farhadi, YOLOv3: An Incremental Improvement, ArXiv.org. (2018). https://arxiv.org/abs/1804.02767v1 (accessed December 9, 2020).

[36] A. Howard, M. Sandler, B. Chen, W. Wang, L.-C. Chen, M. Tan, et al., Searching for MobileNetV3, 2019 IEEE/CVF International Conference on Computer Vision (ICCV). (2019). doi:10.1109/iccv.2019.00140.

[37] X. Sun, B. Xiao, F. Wei, S. Liang, Y. Wei, Integral Human Pose Regression, Proceedings of the European Conference on Computer Vision (ECCV), 529-545.(2017)

[38] S. Ioffe, C. Szegedy, Batch normalization: accelerating deep network training by reducing internal covariate shift, (2015). https://dl.acm.org/doi/10.5555/3045118.3045167 (accessed December 13, 2020).

[39] K. He, X. Zhang, S. Ren, J. Sun, Deep Residual Learning for Image Recognition, 2016 IEEE Conference on Computer Vision and Pattern Recognition (CVPR). (2016). doi:10.1109/cvpr.2016.90.

[40] J. Hu, L. Shen, S. Albanie, G. Sun, E. Wu, Squeeze-and-Excitation Networks, ArXiv.org. (2019). https://arxiv.org/abs/1709.01507 (accessed December 13, 2020).

[41] D. Hendrycks, T. Dietterich, Benchmarking Neural Network Robustness to Common Corruptions and Perturbations, ArXiv.org. (2019). https://arxiv.org/abs/1903.12261 (accessed August 30, 2020).

[42] A.B. Jung, K. Wada, S. Tanaka, C. Reinder, et al. Imgaug, (2020). https://github.com/aleju/imgaug (accessed December 13, 2020)

[43] diiselrong, Train wheelon a rail 2, (2018). https://youtu.be/6oEkVbhT_T8 (accessed December 13, 2020)

[44] T.-Y. Lin, M. Maire, S. Belongie, J. Hays, P. Perona, D. Ramanan, et al., Microsoft COCO: Common Objects in Context, (2018). https://www.microsoft.com/en-us/research/publication/microsoft-coco-common-objects-in-context/ (accessed December 21, 2020).

[45] T. Asano, N. Katoh, Variants for the Hough transform for line detection, Computational Geometry. 06-04 (1996) 231-252. doi: 10.1016/0925-7721(95)00023-2.

[46] D. Shi, Y. Ye, M. Gillwald, M. Hecht, Designing a lightweight 1D convolutional neural network with Bayesian optimization for wheel flat detection using carbody accelerations, International Journal of Rail Transportation. (2020) 1–31. doi:10.1080/23248378.2020.1795942.

[47] H. Abbasian, J. Park, S. Sharma, S. Rella, Speeding Up Deep Learning Inference Using TensorRT. (2020). https://developer.nvidia.com/blog/speeding-up-deep-learning-inference-using-tensorrt/ (accessed December 13, 2020)

## [48]Annex I

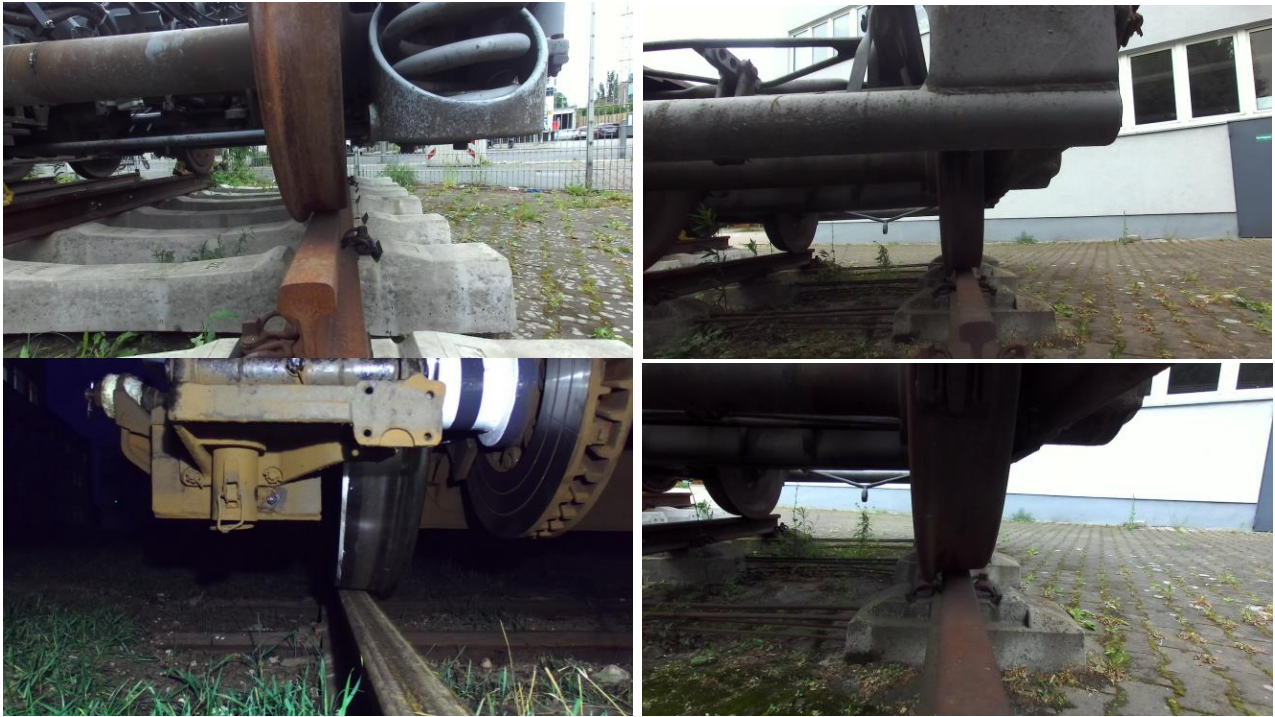

**Fig. 14.** Examples of static images taken on different bogies

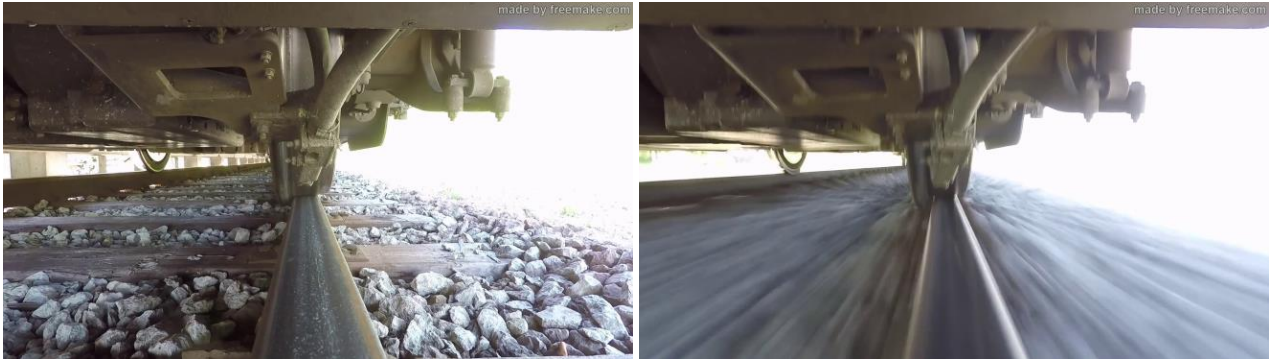

**Fig. 15.** Examples of images from a YouTube video